\ifdefined\pdfminorversion\pdfminorversion=7\fi
\PassOptionsToPackage{numbers,sort&compress,square}{natbib}
\documentclass[11pt,a4paper,onecolumn]{thuc3i}
\usepackage{custom}
\usepackage{xspace}
\usepackage{etoolbox}
\usepackage{caption}

\newcommand{\name}{EgoCaptioner\xspace}

\setheadertext{\name}
\title{\name: Towards Comprehensive and Fine-Grained Egocentric Video Captioning}

\author[1,2]{Yuqi Tang}
\author[3,2,$\dagger$]{Yang Shi}
\author[3]{Yuran Wang}
\author[3]{Tengfei Liu}
\author[2]{Rundong Wang}
\author[1]{Zongyang Qiu}
\author[3]{Qixun Wang}
\author[3]{Zhuoran Zhang}
\author[3]{Xuanyu Zhu}
\author[4]{Yifan Xu}
\author[4]{Yiyan Ji}
\author[5]{Shujie Li}
\author[6]{Shixuan Liu}
\author[6]{Haotian Wang}
\author[2]{Jialu Chen}
\author[2,$\spadesuit$]{Yuanxing Zhang}
\author[2]{Pengfei Wan}
\author[1,$\dagger$]{Chengwei Qin}

\affil[1]{HKUST(GZ)}
\affil[2]{KlingAI Research}
\affil[3]{PKU}
\affil[4]{NJU}
\affil[5]{HKU}
\affil[6]{THU}

\role[\dagger]{Corresponding Author}
\role[\spadesuit]{Project Lead}

\resource{}{%
\href{https://cactusqq.github.io/EgoCaptioner/}{\faHome~Homepage}%
  \hspace{1.5em}%
\href{https://github.com/cactusqq/EgoCaptioner}{\faGithub~Code}%
  \hspace{1.5em}%
\href{https://huggingface.co/Vickyinmyheart824/EgoCaptioner}{\faCube~Model}%
}

\begin{abstract}
Egocentric video captioning requires fine-grained detail and accurate timing, yet current multimodal large language models (MLLMs) often miss brief actions, confuse their order, or misinterpret the wearer's motion.
Existing caption formats also offer limited support for combining detailed narration with precise timing.
We present \name, which pairs a video-level overview of the setting, participants, and the wearer's overall activity with timed steps whose narration adapts to varying perceptual demands and marks key moments at specific timestamps.
We develop a coarse-to-fine annotation pipeline that establishes event structure before expanding step descriptions, yielding 50{,}000 training examples from eight egocentric video sources.
Complementary rewards assess factual accuracy and coverage, measure step-level temporal alignment, and penalize fragmented or redundant narration, guiding caption refinement through Group Relative Policy Optimization (GRPO).
To improve the reliability of factual supervision, we supplement caption references with pose-derived motion facts when sensor data are available.
Across five egocentric benchmarks, \name achieves leading performance among evaluated open-source models in caption quality, caption-conditioned question answering, and temporal alignment.
It also surpasses Gemini 3.1 Pro on PhysCapBench and all evaluated proprietary baselines on Minerva-Ego under the caption-to-QA protocol.
Ablations support the complementary roles of factual and temporal rewards and show that coarse-to-fine annotation provides more effective supervision than single-pass annotation.
\end{abstract}

\usepackage{etoolbox}
\usepackage{caption}

\makeatletter
\renewcommand{\abscontent}{%
    \par\noindent
    \textbf{Abstract $\vert$}\enspace
    {\absfont\theabstract\par}
    \@ifundefined{@keywords}{}{%
        \vskip1em
        \noindent\keywordsfont Keywords: \@keywords
    }%
}

\patchcmd{\maketitle}
    {\vskip8pt}
    {%
        \vskip8pt
        \begin{center}
            \captionsetup{type=figure,font=small,skip=4pt}
            \includegraphics[
                width=0.95\textwidth,
                height=0.30\textheight,
                keepaspectratio
            ]{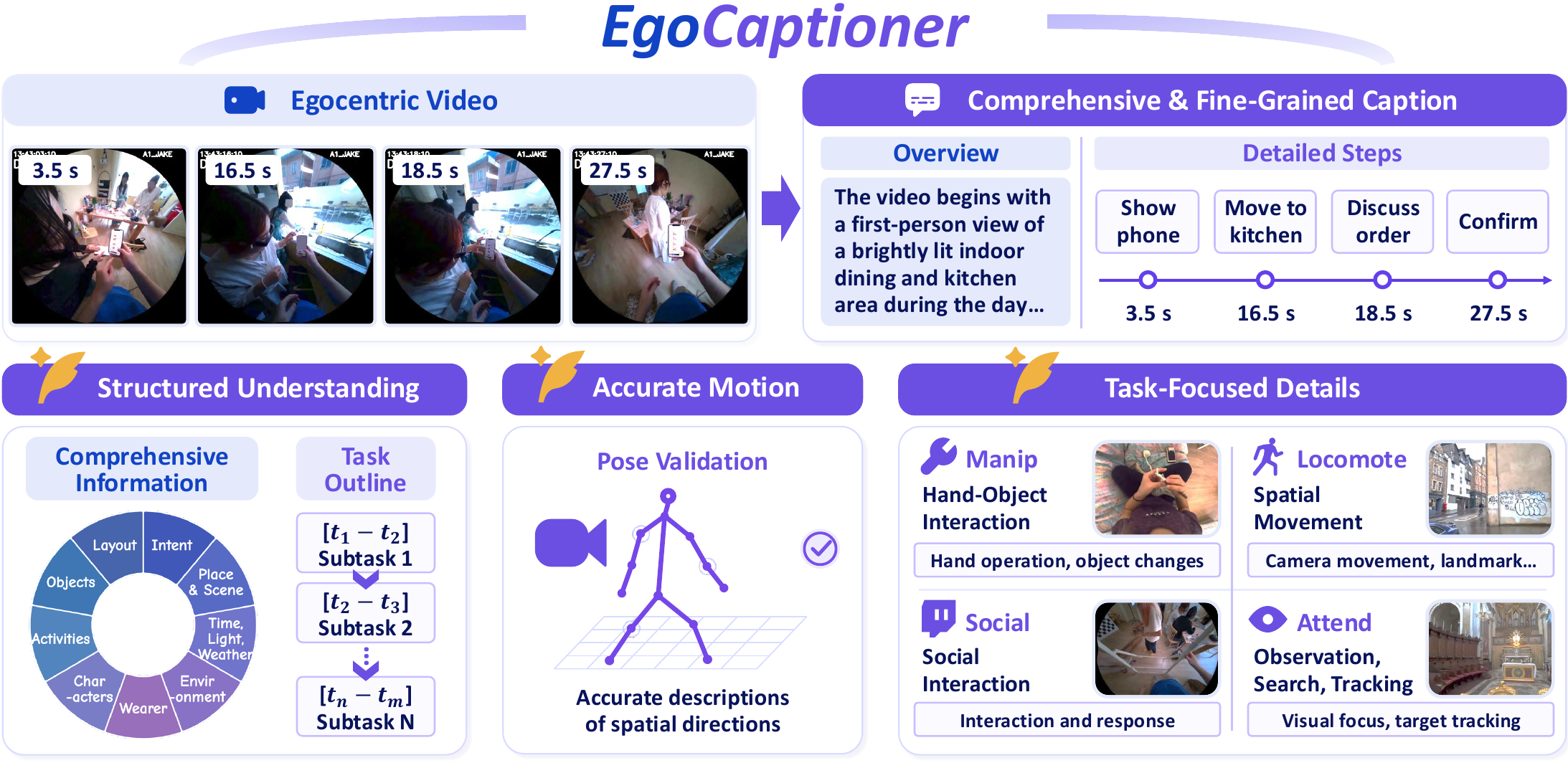}
            \caption{%
                \textbf{Overview of \name.} Given an egocentric video, \name produces a caption with a video-level overview and timestamped steps. The overview establishes the scene context and the wearer's activity and the steps provide temporal structure and narrate details tailored to each activity type. When pose data are available, motion descriptions are verified against sensor measurements.
            }
            \label{fig:teaser}
        \end{center}
        \vspace{4pt}
    }
    {}
    {\PackageError{teaser}{Failed to insert teaser}{}}
\makeatother

\begin{document}
\maketitle


\section{Introduction}
\label{sec:1_introduction}

Egocentric video records how people act in and interact with their surroundings, making it a valuable source of data for embodied AI and video generation~\cite{grauman2022ego4d,yang2025egolife,shi2026mme,leng2026flora,team2026behind}.
Captions make these recordings searchable and allow downstream systems to identify when particular actions occur.
For example, personal assistants retrieve captions as memory to answer questions about the wearer's day~\cite{yang2025egolife}, and robot learning uses timestamped action captions to curate pretraining data~\cite{leng2026flora}. Both applications depend on captions that preserve the details of an activity and associate them with the right moments.

Current multimodal large language models (MLLMs)~\cite{qwen2026qwen38,bytedance2026seed21,shi2025mavors,wang2026beacon,wang2024tarsier,yuan2025tarsier2,shi2026realunify,wang2026monet,zhang2025debiasing,openai2026gpt6astra} can produce detailed video descriptions~\cite{yuan2025tarsier2,zhang2026verus,liu2026refcaptioner,chai2025auroracap,cho2025perceptionlm}.
On egocentric footage, however, they often overlook brief actions, confuse their order, misinterpret the wearer's movements, or omit objects central to an event~\cite{shen2026egoactionbench,plizzari2025egotempo,tashkovska2026soma,nagrani2026minervaego,seth2025egoillusion}.

Part of the problem is that current captioners struggle to decide what to describe and how much detail each activity needs. General captioners \cite{yuan2025tarsier2,chai2025auroracap} cover the visible scene broadly without binding the narrative to the wearer's actions. 
Current egocentric captioners, meanwhile, typically produce a brief narration or specialize in hand--object manipulation~\cite{zhao2023lavila,lu2025egovideo,xu2024egoinstructor,ohkawa2025exo2ego}, leaving activities such as navigation and social interaction less fully described. 
A second challenge is that detail and timing are rarely achieved together. Dense paragraphs often leave unclear when individual actions begin and end, while timestamped captions typically summarize each event in a sentence~\cite{ren2024timechat,huang2024vtimellm} or group details by scene~\cite{yao2026timechatcaptioner}.
Furthermore, the wearer's motion is difficult to interpret from the first-person viewpoint, because the body is often only partially visible, and the captioner must infer travel direction from changes in the scene~\cite{tashkovska2026soma}, cues that can be misleading when head motion diverges from body motion~\cite{matthis2022retinal}.

To address these challenges, we present \name, which describes an egocentric video on two levels (Figure~\ref{fig:teaser}).
A video-level \emph{overview} establishes the setting and spatial layout, introduces the people and objects involved, and summarizes the wearer's activities and any purpose supported by the footage. \emph{Detailed steps} then narrate how the activity unfolds: \name segments the video into temporally bounded events, divides each event into ordered steps with start and end times, and marks \emph{key moments}, such as object state changes, with timestamps.
This two-level structure supports three tasks: full-video captioning, overview captioning, and event-conditioned captioning, in which the model localizes and describes an event named in the instruction.
We train \name{} by supervised fine-tuning on 50{,}000 curated examples from eight egocentric video sources, followed by reinforcement learning with rewards for factual content, temporal alignment, and caption style.

We evaluate \name across several benchmarks on four dimensions: caption quality, caption-conditioned QA, coverage of physical event details, and temporal alignment. Compared with its Qwen3.5-9B backbone, \name improves by 27.6 points in
PhysCapBench~\cite{wang2026physcapbench} Overall, surpassing Gemini 3.1 Pro, and by 16.8 percentage
points in EgoTempo~\cite{plizzari2025egotempo} accuracy. It also achieves the highest PSJ score among open-source models for step segmentation on EgoHome~\cite{perceptron2026egohome}. Our ablations examine how training data scale, annotation format, and reward design each affect performance. Further analysis shows that tailoring descriptive detail to the activity helps captions capture the aspects that human observers find salient.

Our contributions are as follows:
\begin{itemize}
    \item We formulate egocentric captioning as a video overview and timed steps that captures task-focused details, action intervals, and key moments.
    \item We develop a coarse-to-fine annotation pipeline and use it to construct 50{,}000 training examples from eight egocentric video sources.
    \item We train \name{} with rewards for factual content, timing, and style, achieving the best open-source results on five benchmarks.
\end{itemize}

\section{Related Work}
\label{sec:related_work}

\paragraph{Multimodal Large Language Models}
Recent multimodal large language models (MLLMs) advance temporal understanding through scaled synthetic supervision~\cite{zhang2024llavavideo}, multi-granularity descriptions~\cite{wang2024tarsier}, and time-aware position encoding~\cite{bai2025qwen25vl,bai2025qwen3vl}.
Yet egocentric benchmarks reveal persistent hallucinations about object interactions~\cite{seth2025egoillusion} and temporal reasoning errors~\cite{plizzari2025egotempo}, calling for descriptions that faithfully ground wearer actions and camera motion in the video's temporal structure.

\paragraph{Egocentric Video Understanding}
First-person video interleaves rapid head motion with hand--object manipulation, making the wearer largely invisible and scene dynamics viewpoint-dependent.
Representation methods address this by grouping spatiotemporal tokens into scene entities~\cite{han2024henasy} or learning hand--object dynamics with motion adapters~\cite{lu2025egovideo}, while captioning approaches leverage third-person references~\cite{xu2024egoinstructor}, cross-view transfer~\cite{ohkawa2025exo2ego}, and recursive composition across temporal scales~\cite{islam2024videocap}.
Such captions further serve as retrievable memory for question answering~\cite{yang2025egolife} and as timestamped supervision for robot pretraining~\cite{leng2026flora}.
We produce a video-level overview with timestamped steps and key moments, emphasizing hand roles, motion, and task-focused details.

\section{Method}
\label{sec:3_method}

\subsection{Task Formulation}
\label{sec:task_form}

We describe an egocentric video at two levels.
A video-level \emph{overview} establishes the setting and summarizes what the camera wearer is doing.
\emph{Steps} then narrate how the activity unfolds, organized by timed \emph{events} that cover the video in consecutive, non-overlapping intervals.

\paragraph{Overview.}
The overview sets the scene by describing the location, surroundings, lighting and the time of day and weather.
Within this setting, it describes the wearer, other participants, and relevant objects, noting their positions relative to one another.
Drawing on this context and the actions shown, it also summarizes what the wearer is doing overall and, when the goal is apparent, what they are trying to accomplish.

\paragraph{Steps.}
We segment each video into non-overlapping events, each covering a coherent activity, and divide each event into ordered steps with start and end times.
The steps narrate what happens within the event, emphasizing the details that matter for the activity.
A step may also record a \emph{key moment} in a sentence beginning with \texttt{At [t]}, marking the instant at which something relevant happens.

\paragraph{Tasks.}
The two levels define three instruction-following tasks.
(1) \emph{Full-video captioning} produces the overview followed by the steps of all events.
(2) \emph{Overview captioning} produces the overview alone.
(3) \emph{Event-conditioned captioning} produces the steps of an event named in the instruction, requiring the model to localize the event before describing it.

\subsection{Data Construction}
\label{sec:data_construction}

\begin{figure}[t]
    \centering
    \includegraphics[width=\linewidth]{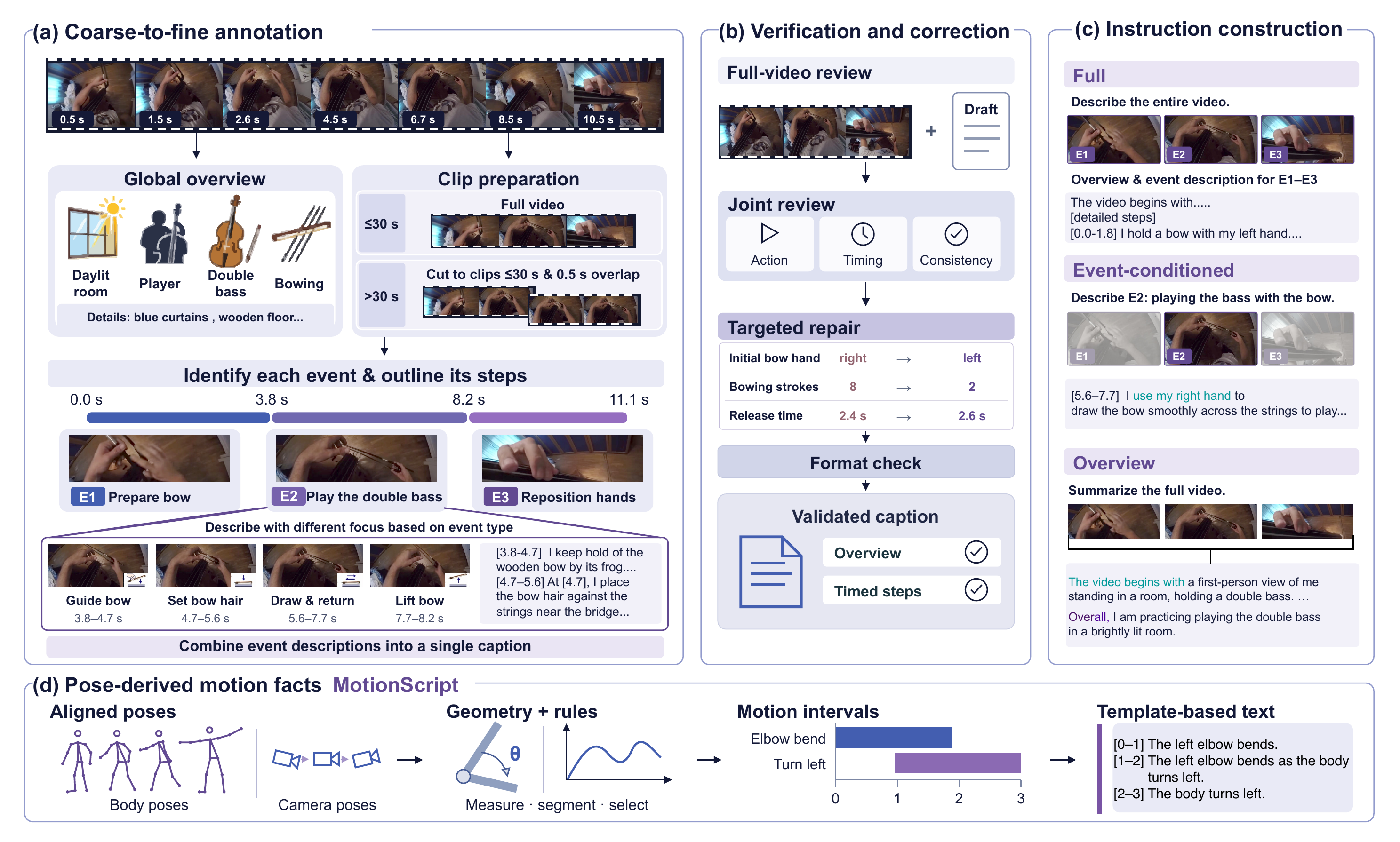}
    \caption{\textbf{Data construction pipeline.} (a) Gemini-3.1-Pro and Seed2.1 pro annotate each video coarse to fine, from a global overview and event outlines to timestamped steps. (b) The draft caption is reviewed against the full video and corrected. (c) Each validated caption can yield full-video, event-conditioned, and overview instructions. (d) Body and camera poses are converted into timed motion facts.}
    \label{fig:data_construction}
\end{figure}

We collect egocentric videos spanning daily activities, procedural tasks, outdoor walking, and manipulation from Ego4D Goal-Step~\cite{song2023goalstep}, the Ego4D subset of EgoHOD~\cite{lu2025egovideo}, Sekai~\cite{li2025sekai}, EgoVid-5M~\cite{wang2024egovid}, the EgoPro subset of EgoSuite-Open100K~\cite{lightwheel2026egosuite}, HumanPlus-1000~\cite{humanplus1000}, the processed release of RekaDaily-10k \cite{reka2026rekadaily} and EgoLife~\cite{yang2025egolife}.
We build the annotation coarse-to-fine with two complementary MLLMs, Gemini-3.1-Pro~\cite{google2026gemini31pro} and Seed2.1 pro~\cite{bytedance2026seed21}.

\paragraph{Task-focused narration.}
Different activities call for different details, so we define four event types and use each type's narration emphasis to prompt the annotator, steering the description toward the details most relevant to that activity.
Hand--object manipulation (\textsc{manip}) centers on what the wearer does to an object, each hand's role, how the object changes, and where it is placed afterward.
Locomotion (\textsc{locomote}) records the travel mode, direction, and route relative to landmarks, using camera motion cues such as bobbing or sway to identify the mode.
Social interaction (\textsc{social}) covers who participates, how they take turns acting and responding, and what gestures or expressions clarify intent.
Visual attention (\textsc{attend}) tracks what the wearer observes, how attention shifts between targets, and when a target becomes occluded or reappears.

Within a step, \emph{key moments} mark sudden transitions, such as an object state change, a mishap, or a person entering or leaving the scene.
Each is written as \texttt{At [t],} followed by what happens at that instant.

\paragraph{Coarse-to-fine annotation.}
The annotation proceeds in two stages (Figure~\ref{fig:data_construction}(a)).
\emph{(i) Structure.} Gemini writes the overview from the full video. Seed2.1 pro identifies the events, their temporal boundaries and types, and an outline of steps for each event.
\emph{(ii) Detail.} Gemini expands each outline into timestamped step descriptions with key moments, prompted with the narration emphasis of the event's type. It views an event crop with up to 1\,s of surrounding context on each side.
Videos longer than 30\,s are annotated in clips of at most 30\,s with 0.5\,s overlap, and Gemini merges the clip annotations into a single timeline.
The two stages produce a complete draft caption for verification.

\paragraph{Verification and correction.}
Seed2.1 pro then reviews the draft against the full video (Figure~\ref{fig:data_construction}(b)), jointly checking the described actions, their timing, and the consistency of the events with one another.
It corrects localized errors in place, and a final format check yields the validated caption.

From each validated annotation we construct instructions for the three tasks in Sec.~\ref{sec:task_form} (Figure~\ref{fig:data_construction}(c)).
For event-conditioned examples we only provide the event's name without its time localization.

\paragraph{Pose-derived motion facts.}
For videos with pose or trajectory recordings, we convert sensor measurements into discrete motion facts following MotionScript~\cite{yazdian2025motionscript}, each describing the wearer's body movement or a change in head-mounted viewing direction together with its temporal span.
Because these facts come from sensors, they provide motion evidence independent of the reference captions and serve as additional reference facts during reinforcement learning (Sec.~\ref{sec:rl}).

\subsection{Supervised Fine-Tuning}
\label{sec:sft}
\begingroup
\emergencystretch=1em
We initialize our captioning policy by supervised fine-tuning (SFT) on 50{,}000 curated video--instruction--caption triplets, with videos up to 120\,s long.
The training data comprise an equal mixture of full-video, overview, and event-conditioned captioning examples.
This stage trains the model to produce egocentric descriptions at the requested granularity and provides the initial policy for subsequent reinforcement learning.
\par\endgroup

\subsection{Reinforcement Learning}
\label{sec:rl}

\begin{figure}[t]
    \centering
    \includegraphics[width=\linewidth]{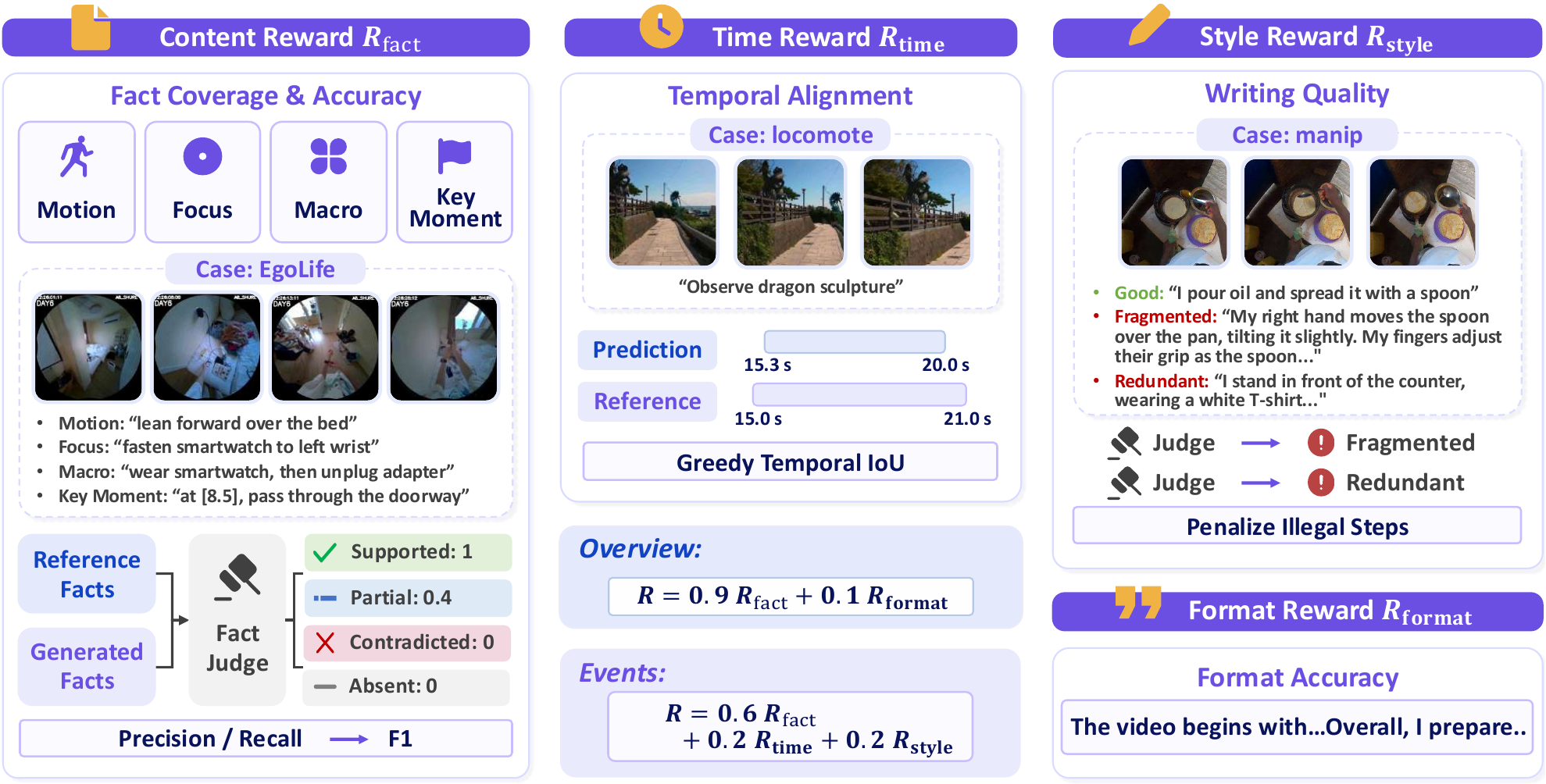}
    \caption{\textbf{Illustration of the rewards designed for egocentric captioning.} Fact, time, and style rewards are combined for step-level captions, while overview captions pair the content reward with a format reward. The content reward $R_{\mathrm{fact}}$ is the F1 of judge-scored precision and recall over extracted facts, the time reward $R_{\mathrm{time}}$ greedily matches predicted and reference steps by temporal IoU, and the style reward $R_{\mathrm{style}}$ penalizes steps that judges flag as fragmented or redundant.}
    \label{fig:reward_design}
\end{figure}

We refine the SFT policy $\pi_\theta$ with Group Relative Policy Optimization (GRPO)~\cite{shao2024grpo}.
Given a video $v$ and a task instruction $q$, let $x=(v,q)$, $\hat y\sim\pi_\theta(\cdot\mid x)$ denote a sampled caption, and $y^\star$ its reference.
As shown in Figure~\ref{fig:reward_design}, we guide the optimization with three complementary rewards.
A fact-based reward measures how faithfully and completely the caption describes the video.
For steps, a temporal alignment reward measures how well the predicted step boundaries and granularity match the reference, and a style reward discourages fragmented or redundant writing within each step.

\paragraph{Fact-based reward.}
We evaluate captions using four fact types defined by the caption structure in Sec.~\ref{sec:task_form}.
\emph{Motion} facts describe the wearer's body movement and posture, as well as changes or stability in the head-mounted view. When pose data are available, we also include the pose-derived facts from Sec.~\ref{sec:data_construction}.
\emph{Focus} facts capture the task-focused details defined for each event type in Sec.~\ref{sec:data_construction}.
\emph{Macro} facts describe the scene, the visual style and the wearer's overall activity and purpose.
\emph{Key-moment} facts, pair a timestamp with the change occurring at that instant and count as correct only when both parts are.

We extract the facts stated in each caption and write $\mathcal F(\hat y)$ and $\mathcal F(y^\star)$ for the predicted and reference fact sets.
Gemini-3.1-Pro~\cite{google2026gemini31pro} then judges each fact against the opposite caption, assigning a score $s(f\mid y)\in[0,1]$ with partial credit when the core of a fact is correct but some details are missing.
Precision averages these scores over $\mathcal F(\hat y)$ and recall over $\mathcal F(y^\star)$. The reward is their F1:
\begin{equation}
R_{\mathrm{fact}}=\frac{2pr}{p+r}.
\label{eq:fact_reward}
\end{equation}

\paragraph{Temporal alignment reward.}
The temporal reward measures agreement with the reference in both the placement of step boundaries and the granularity of the segmentation. We greedily match predicted intervals $\widehat{\mathcal I}$ to reference intervals $\mathcal I$ in descending order of temporal intersection over union (tIoU), using each interval at most once. We sum the overlaps of the matched pairs $\Pi$ and normalize by the larger of the two step counts, so that unmatched steps contribute zero:
\begin{equation}
    R_{\mathrm{time}}
    = \frac{\sum_{(i,j)\in\Pi}\operatorname{tIoU}(\widehat I_i,I_j)}
           {\max\bigl(|\widehat{\mathcal I}|,|\mathcal I|\bigr)}.
    \label{eq:time_reward}
\end{equation}

\paragraph{Style reward.}
Even a factually accurate, well-timed caption can leave the action unclear or repeat information the reader already has.
$R_{\mathrm{style}}$ captures these two failures and calibrates the writing toward fluent, natural captions.
Two independent judges read the full caption: one flags \emph{fragmented} steps, which are hand--object interaction steps that describe only hand movements or contact without specifying what the camera wearer does to the object, and the other flags \emph{redundant} steps, which contain a sentence that only restates background given earlier.
With $\mathcal F$ the union of flagged steps and $n_s$ the number of steps,
\begin{equation}
    R_{\mathrm{style}}=1-\frac{|\mathcal F|}{n_s}.
    \label{eq:style_reward}
\end{equation}

\paragraph{Reward aggregation.}
We assign $R=-1$ to captions that are truncated, degenerate into repetition, or have nonpositive duration.
For valid rollouts in event-conditioned and full-video captioning, we compute
\begin{equation}
    R = 0.6R_{\mathrm{fact}} + 0.2R_{\mathrm{time}} + 0.2R_{\mathrm{style}}.
    \label{eq:reward_aggregation}
\end{equation}
For valid overview captioning, we use $R = 0.9R_{\mathrm{fact}} + 0.1R_{\mathrm{format}}$, where $R_{\mathrm{format}}$ rewards overviews that open with a description of the scene and close with a summary of the wearer's activity.

\paragraph{Training objective.}
For each input $x$, GRPO samples a group of $G$ captions $\{\hat y_i\}_{i=1}^{G}$ from the rollout policy $\pi_{\mathrm{gen}}$.
The reward $R_i$ of each caption is standardized within the group, $A_i=(R_i-\mu)/(\sigma+\epsilon_0)$, where $\mu$ and $\sigma$ are the group mean and standard deviation and $\epsilon_0$ is a small constant.
Let $\rho_{i,t}$ be the probability ratio between the current policy $\pi_\theta$ and the pre-update policy $\pi_{\theta_{\mathrm{old}}}$ for the $t$-th token of $\hat y_i$, and $\bar\rho_{i,t}$ its clipped value in $[1-\varepsilon,1+\varepsilon]$.
We maximize
\begin{equation}
    \mathcal J(\theta)
    =\mathbb E\!\left[
      \frac{\sum_{i,t}\min(\rho_{i,t}A_i,\bar\rho_{i,t}A_i)}
           {\sum_i|\hat y_i|}\right]
    -\beta\,\mathbb D_{\mathrm{KL}}(\pi_\theta\|\pi_{\mathrm{SFT}}),
    \label{eq:grpo_objective}
\end{equation}
which averages the clipped objective over all tokens in the group, and $\beta$ weights a KL penalty toward the frozen SFT policy.
The remaining reward details and optimization settings are provided in the supplementary material.

\section{Experiments}
\label{sec:4_experiment}

\providecommand{\tbd}{\textcolor{red}{XX}}
\providecommand{\pendingresult}{\phantom{00.00}}
\providecommand{\best}[1]{\textbf{#1}}
\providecommand{\second}[1]{\underline{#1}}
\providecommand{\up}{$\uparrow$}
\providecommand{\down}{$\downarrow$}

To comprehensively evaluate the ability of \name to produce detailed, time-aware and task-focused descriptions, we investigate three research questions (RQs): \textbf{RQ1:} How well does it understand egocentric videos? \textbf{RQ2:} Which design choices in our annotation and training pipeline contribute to the performance? \textbf{RQ3:} Beyond benchmark scores, how well do its descriptions reflect the detailed, task-focused design in practice?

\begin{table}[!t]
\centering
\small
\caption{\textbf{Caption quality, video understanding, and temporal alignment (RQ1).}
All scores are scaled by 100 except EgoTempo Score (0--5).
B@4 and R-L denote BLEU-4 and ROUGE-L; PhysCapBench Overall averages seven dimensions.
$\dagger$ denotes direct video QA results.
\colorbox{bestgreen}{\strut Best} and \underline{second-best} indicate rankings among open-source models.}
\label{tab:main_results}
\footnotesize
\setlength{\tabcolsep}{1.5pt}
\renewcommand{\arraystretch}{1.08}
\resizebox{\linewidth}{!}{%
\begin{tabular}{@{}l*{12}{c}@{}}
\toprule
 & \multicolumn{4}{c}{\textbf{EgoMe-Overview}}
 & \multicolumn{3}{c}{\textbf{EgoMe-Steps}}
 & \multicolumn{2}{c}{\textbf{EgoTempo}}
 & \multicolumn{1}{c}{\textbf{Minerva-Ego}}
 & \multicolumn{1}{c}{\textbf{EgoHome}}
 & \multicolumn{1}{c}{\textbf{PhysCapBench}} \\
\cmidrule(lr){2-5}
\cmidrule(lr){6-8}
\cmidrule(lr){9-10}
\cmidrule(lr){11-11}
\cmidrule(lr){12-12}
\cmidrule(l){13-13}
\textbf{Model}
& B@4\up & METEOR\up & R-L\up & CIDEr\up
& METEOR\up & SODA-c\up & SODA-d\up
& Acc\up & Score\up & Acc$_{\rm open}$\up
& PSJ\up & Overall\up \\
\midrule

\rowcolor{groupblue}
\multicolumn{13}{c}{\textbf{Proprietary MLLMs}} \\
Gemini 3.1 Pro & 3.32 & 18.62 & 24.45 & 5.61 & 3.98 & 7.41 & 46.80 & 23.60 & 1.444 & 4.29 & 52.86 & 75.06 \\
Gemini 3.8 Flash & 2.62 & 19.27 & 22.12 & 0.79 & 6.26 & 10.34 & 48.31 & 41.20 & 2.310 & 7.14 & 71.09 & 82.77 \\
GPT-6-Astra & 1.28 & 16.12 & 15.95 & 0.00 & 5.91 & 6.25 & 43.12 & 52.60 & 2.842 & 15.71 & 79.42 & 84.87 \\

\midrule
\rowcolor{groupblue}
\multicolumn{13}{c}{\textbf{Open-source MLLMs and Captioners}} \\
Qwen3-VL-8B-Instruct & 1.59 & 15.72 & 16.50 & 0.05 & 5.73 & 6.34 & 43.33 & 14.00 & 0.966 & 1.43 & 55.88 & 68.94 \\
Qwen3-VL-8B-Thinking & 1.73 & 15.93 & 18.18 & 0.45 & 4.68 & 5.48 & 38.05 & 17.20 & 1.100 & 1.43 & 54.02 & 68.98 \\
Qwen3.5-9B & 2.65 & \second{17.58} & 22.59 & \second{4.01} & 5.71 & \second{8.02} & \second{50.25} & 21.60 & 1.388 & 0.00 & 46.90 & 51.24 \\
Qwen3.8-27B & 1.37 & 15.71 & 14.82 & 0.02 & \second{7.24} & 7.09 & 45.67 & 30.80 & 1.842 & 11.43 & \second{69.28} & 69.55 \\
Molmo2-8B & 1.20 & 14.64 & 15.02 & 0.11 & 4.76 & 4.30 & 38.19 & 19.20 & 1.204 & 7.14 & 54.20 & 53.20 \\
Tarsier2-7B & 2.72 & 17.36 & 21.05 & 1.01 & 0.00 & 0.00 & 0.00 & 18.00 & 1.212 & 2.86 & 0.00 & 62.92 \\
CapRL-Video-4B & 1.34 & 14.23 & 12.50 & 0.00 & 4.10 & 3.17 & 30.25 & \second{31.80} & \second{1.978} & 10.00 & 62.87 & \second{75.94} \\
TimeLens-8B & 1.59 & 15.35 & 16.23 & 0.10 & 6.37 & 7.10 & 46.93 & 17.80 & 1.124 & 1.43 & 60.15 & 60.25 \\
AVoCaDO & 1.14 & 13.53 & 12.64 & 0.00 & 0.00 & 0.00 & 0.00 & 13.60 & 0.918 & 4.29 & 0.00 & 60.84 \\

\midrule
\rowcolor{groupblue}
\multicolumn{13}{c}{\textbf{Egocentric MLLMs and Captioners}} \\
EgoThinker$^\dagger$ & \second{2.96} & 15.86 & \second{23.03} & \best{4.85} & 1.86 & 1.79 & 14.93 & 24.40 & 1.770 & \second{15.71} & 9.48 & 39.88 \\
EgoGPT & 2.01 & 14.17 & 19.98 & 2.50 & 0.00 & 0.00 & 0.00 & 9.20 & 0.690 & 0.00 & 0.00 & 32.92 \\

\midrule
\textbf{\name} & \best{4.03} & \best{20.40} & \best{23.59} & 0.65 & \best{12.26} & \best{8.04} & \best{56.38} & \best{38.40} & \best{2.066} & \best{18.57} & \best{75.51} & \best{78.84} \\
\bottomrule
\end{tabular}%
}
\end{table}

\subsection{Experimental Setup}
\label{sec:exp_setup}

\paragraph{Data and training.}
We initialize \name from Qwen3.5-9B~\cite{qwen35blog} and fine-tune it with LoRA~\cite{hu2022lora} (rank 32) on the data described in Sec.~\ref{sec:sft} for 2 epochs with a learning rate of $1{\times}10^{-4}$.
We then apply GRPO on 8{,}000 video--instruction pairs balanced across the three captioning tasks, sampling $G{=}8$ rollouts per input with a batch size of 32 and a learning rate of $1{\times}10^{-5}$.
Both stages use 16 GPUs.

\paragraph{Baselines.}
Table~\ref{tab:main_results} compares \name with three groups:
(i) proprietary MLLMs (Gemini 3.1 Pro~\cite{google2026gemini31pro}, Gemini 3.8 Flash~\cite{deepmind2026gemini38}, GPT-6-Astra~\cite{openai2026gpt6astra}),
(ii) open-source MLLMs and captioners (Qwen3-VL-8B~\cite{bai2025qwen3vl}, Qwen3.5-9B~\cite{qwen35blog} and Qwen3.8-27B~\cite{qwen2026qwen38}, Molmo2-8B~\cite{clark2026molmo2}, TimeLens-8B~\cite{zhang2025timelens}, Tarsier2-7B~\cite{yuan2025tarsier2}, CapRL-Video-4B~\cite{yang2026caprlpp}, AVoCaDO~\cite{chen2026avocado}), and
(iii) egocentric MLLMs (EgoThinker~\cite{pei2025egothinker}, EgoGPT~\cite{yang2025egolife}).

\paragraph{Benchmarks and metrics.}
For RQ1, we evaluate on five egocentric benchmarks: EgoMe~\cite{qiu2025egome}, EgoHome Subtask~\cite{perceptron2026egohome}, and PhysCapBench~\cite{wang2026physcapbench} under their official metrics, and Minerva-Ego~\cite{nagrani2026minervaego} (videos up to 120\,s) and EgoTempo~\cite{plizzari2025egotempo}, two video QA benchmarks that we adapt to a caption-to-QA protocol, where Gemini 3.8 Flash answers each question using only the generated caption and the question. A separate call to the same model judges the answer against the reference.

\paragraph{Inference.}
We uniformly sample frames across the entire video at a target rate of
4\,fps, capped at 240 frames. For longer videos, the effective sampling
rate is reduced to preserve full temporal coverage.
Within each protocol, we keep benchmark instructions and scoring fixed across models.
Further details are provided in the supplementary material.

\subsection{RQ1: Egocentric Video Understanding}
\label{sec:main_results}

We first assess how well \name understands egocentric videos (Table~\ref{tab:main_results}).
The benchmarks probe two sides of this capability: how faithful and informative the descriptions are, measured by caption quality against references and by detail-oriented QA, and whether the described events are anchored to the correct moments.

\paragraph{Description quality and video QA.}
\name delivers the strongest open-source results on EgoMe, EgoTempo, and Minerva-Ego.
On EgoMe-Overview it reaches the highest BLEU-4 and METEOR among all models including proprietary ones, and on EgoMe-Steps it likewise leads in METEOR and SODA-d.
EgoTempo and Minerva-Ego probe fine-grained facts such as manipulated objects, individual actions, and their order, so a text-only reader can answer only when the caption records those facts.
Reading from \name's captions yields the best open-source accuracy on both benchmarks, and on Minerva-Ego it surpasses all proprietary models.
On PhysCapBench, \name scores 78.84 Overall, the best among open-source models and above Gemini 3.1 Pro.

\paragraph{Temporal alignment.}
On EgoHome, we compare our predicted steps with human-annotated subtask timelines for household activities.
The permissible subsegment Jaccard (PSJ) score measures the fraction of time for which the two timelines agree on the action being performed.
The judge can match a step in either timeline to several finer steps in the other if their descriptions refer to the same underlying action.
\name scores 75.51 on PSJ, ranking first among the open-source models evaluated and outperforming Gemini 3.8 Flash.
Tarsier2, AVoCaDO, and EgoGPT score zero on all step-level metrics because they rarely produce parseable timestamps.

Taken together, \name is competitive with proprietary models across all five benchmarks while being the strongest open-source model on every one of them.

\subsection{RQ2: Which Design Choices Are Effective?}
\label{sec:ablation}

To understand the gains in RQ1, we examine the contributions of the annotation procedure, reward design, and amount of training data.

\paragraph{Training stages, data scale, and rewards.}
Table~\ref{tab:ablation_rewards} reports three groups of comparisons.
The first group compares the base model, SFT, and SFT followed by GRPO to show the contribution of each training stage.
The second varies the amount of SFT data (10k, 25k, 50k triplets, sampled by video with the task mixture preserved) under the same recipe.
The third removes one GRPO reward term at a time from the full objective, keeping SFT initialization and data fixed.

On our sampled subsets, SFT alone already brings large PSJ gains over the base model: even 10k triplets recover most of the PSJ improvement (67.79 vs.\ 71.63 at 50k), while METEOR and SODA-d improve steadily as the data scale increases to 50k.
GRPO then lifts METEOR, SODA-d, and PSJ further.
Without $R_{\mathrm{fact}}$, all metrics fall below the SFT initialization, as the remaining rewards supervise timing and style but place no constraint on content.
Without $R_{\mathrm{time}}$, PSJ falls below SFT while METEOR is unaffected, confirming that content quality alone does not produce aligned timestamps.
Without $R_{\mathrm{style}}$, SODA-d and PSJ both decrease relative to the full objective.
Notably, PSJ responds to all three ablations: it credits a step only when its description matches the video \emph{and} its interval aligns with the reference, PSJ reflects both content and temporal fidelity.

\begin{table}[t]
\centering
\footnotesize
\caption{\textbf{Training-stage, data-scale, and reward ablations (RQ2).}
METEOR and SODA-d are on EgoMe Overview and Steps; PSJ is on EgoHome.
The SFT rows vary only the amount of SFT data under the same recipe.
Fact reward is ablated across all three training tasks, whereas temporal and style rewards are ablated only for full-video and event-conditioned captioning.}
\label{tab:ablation_rewards}
\setlength{\tabcolsep}{2pt}
\begin{tabular*}{\columnwidth}{@{\extracolsep{\fill}}l*{3}{c}@{}}
\toprule
\textbf{Variant} & \textbf{METEOR}$\uparrow$ & \textbf{SODA-d}$\uparrow$ & \textbf{PSJ}$\uparrow$ \\
\midrule
Base & 17.58 & 50.25 & 46.90 \\
SFT (10k) & 16.06 & 42.20 & 67.79 \\
SFT (25k) & 16.37 & 43.80 & 68.63 \\
SFT (50k) & 19.91 & 53.25 & 71.63 \\
\midrule
w/o $R_{\mathrm{fact}}$ & 18.93 & 51.05 & 67.97 \\
w/o $R_{\mathrm{time}}$ & 20.17 & 53.50 & 66.81 \\
w/o $R_{\mathrm{style}}$ & \best{20.41} & 55.23 & 73.22 \\
\textbf{Full} & 20.40 & \best{56.38} & \best{75.51} \\
\bottomrule
\end{tabular*}
\end{table}

\paragraph{Coarse-to-fine annotation.}
To isolate the effect of our multi-stage annotation pipeline on data quality, we construct a single-pass baseline: we prompt Gemini 3.1 Pro to generate a complete caption in our target format from a single call per video, bypassing the intermediate event segmentation and per-event description stages.
The two conditions share the same source videos, training volume, output format, and training hyperparameters; the single-pass prompt is detailed in the supplementary material.
As shown in Table~\ref{tab:ablation_annotation}, the coarse-to-fine supervision improves every metric.
The largest gains appear on detail-oriented QA, where EgoTempo accuracy rises by nearly 10 points and Minerva-Ego more than doubles, and on temporal alignment, where PSJ rises by about 9 points.
In the single-pass condition the annotator segments and describes each step in one reading, producing coarser boundaries and fewer step-level facts.
Our procedure fixes the event structure first and then describes each step from its event crop, which yields both aligned boundaries and the fine-grained facts.

\begin{table}[t]
\centering
\footnotesize
\caption{\textbf{Annotation-procedure ablation after SFT (RQ2).}
Both variants use matched training videos and data volume.}
\label{tab:ablation_annotation}
\setlength{\tabcolsep}{2pt}
\begin{tabular*}{\columnwidth}{@{\extracolsep{\fill}}l*{5}{c}@{}}
\toprule
 & \textbf{Overview} & \textbf{Steps} & \textbf{EgoTempo} & \textbf{Minerva} & \textbf{EgoHome} \\
\textbf{Method} & METEOR$\uparrow$ & SODA-d$\uparrow$ & Acc$\uparrow$ & Acc$\uparrow$ & PSJ$\uparrow$ \\
\midrule
Single-pass & 17.57 & 48.43 & 26.60 & 7.14 & 62.89 \\
\textbf{Ours} & 19.91 & 53.25 & 36.40 & 18.57 & 71.63 \\
\bottomrule
\end{tabular*}
\end{table}

\begin{figure}[t]
\centering
\IfFileExists{Fig/cases.pdf}{%
  \includegraphics[width=\linewidth]{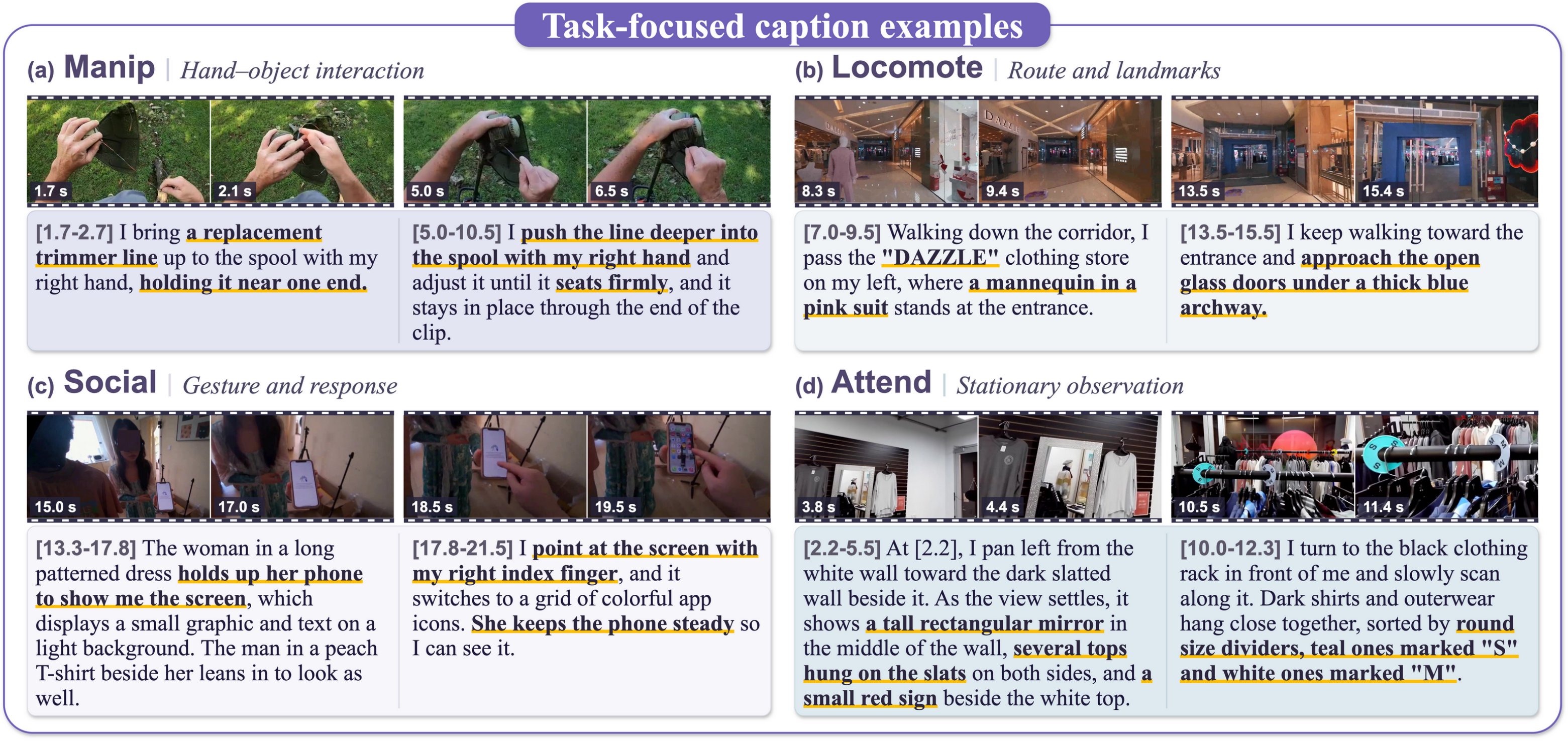}%
}{%
  \input{Fig/rq_activity_cases}%
}
\caption{\textbf{Task-focused detail and temporal granularity (RQ3).}
Each panel shows \name's timed step descriptions for one activity type, with task-relevant details highlighted.}
\label{fig:activity_cases}
\end{figure}

\subsection{RQ3: Detailed and Task-Focused in Practice?}
\label{sec:analysis}

We examine two questions about what the captions contain in practice: whether the activity-type narration emphasis produces different details for different activities, and whether the quality gains come from denser information per token rather than simply longer output.

\paragraph{Task-focused detail.}
Figure~\ref{fig:activity_cases} illustrates how \name shifts its descriptive focus to match the wearer's activity.
For manipulation, the caption decomposes a continuous action into atomic steps, preserving the purpose and target of each motion.
In locomotion, the focus turns to identifiable landmarks and spatial cues that let a reader trace the wearer's route through the scene.
A social exchange requires tracking both parties, so the caption records gestures and responses from each side of the interaction.
When the wearer pauses to observe, the description follows the direction and sequence of their gaze, picking out fine-grained visual details of the objects inspected.

\begin{figure}[htbp]
\centering
\includegraphics[width=0.70\linewidth]{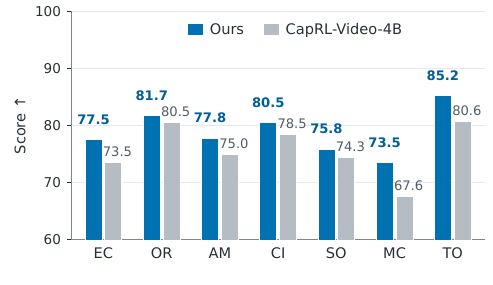}
\vspace{-2em}
\caption{\textbf{Per-dimension PhysCapBench scores (RQ3).}
\name vs.\ CapRL-Video-4B at comparable caption length.
EC: event coverage, OR: objects and roles, AM: actions and motion, CI: contact and interaction, SO: states and outcomes, MC: physical mechanisms and causality, TO: temporal order.}
\label{fig:physcap_dims}
\end{figure}

\paragraph{Information density at comparable length.}
We compare \name with CapRL-Video-4B, the strongest captioner baseline on the egocentric subset of PhysCapBench, across all seven evaluation dimensions (Figure~\ref{fig:physcap_dims}).
The two models produce captions of comparable mean length (both 548 words per video), so differences reflect the information each word carries rather than sheer verbosity.
At this length, \name leads on every dimension, with the largest gains in physical mechanisms and causality (+5.88 points) and temporal order (+4.57 points).
For a reader reconstructing a procedure from text, knowing which actions came first and how they caused the observed changes is exactly what separates a usable caption from a list of observations.
This suggests that \name's captions can serve as a practical text interface to egocentric video, enabling applications such as step-by-step instruction generation and activity summarization.

\section{Conclusion}
\label{sec:5_conclusion}

We presented \name, an egocentric video captioner trained with coarse-to-fine supervision and reinforcement learning that jointly optimizes factual content, temporal alignment, and caption style.
\name achieves the strongest open-source results on five egocentric benchmarks and is competitive with proprietary models across caption quality, detail-oriented QA, and temporal alignment.
Ablations show that coarse-to-fine annotation, which establishes event structure before step descriptions, and joint optimization of factual and temporal rewards are both necessary for consistent gains.
Per-dimension analysis on PhysCapBench further reveals that task-focused narration contributes most where understanding causality and temporal order is required, not merely naming visible objects.

\begingroup
\setlength{\bibsep}{3pt plus 1pt minus 1pt}
\bibliographystyle{plainnat}
\bibliography{ref}
\endgroup

\clearpage
\appendix
\section*{Supplementary Material}

\section{Qualitative Examples}
\label{sec:supp_cases}

\subsection{\name Captions across Activity Types}
\label{sec:supp_cases_model}
\begin{figure}[!htp]
\centering
\includegraphics[width=\textwidth,height=0.85\textheight,keepaspectratio]{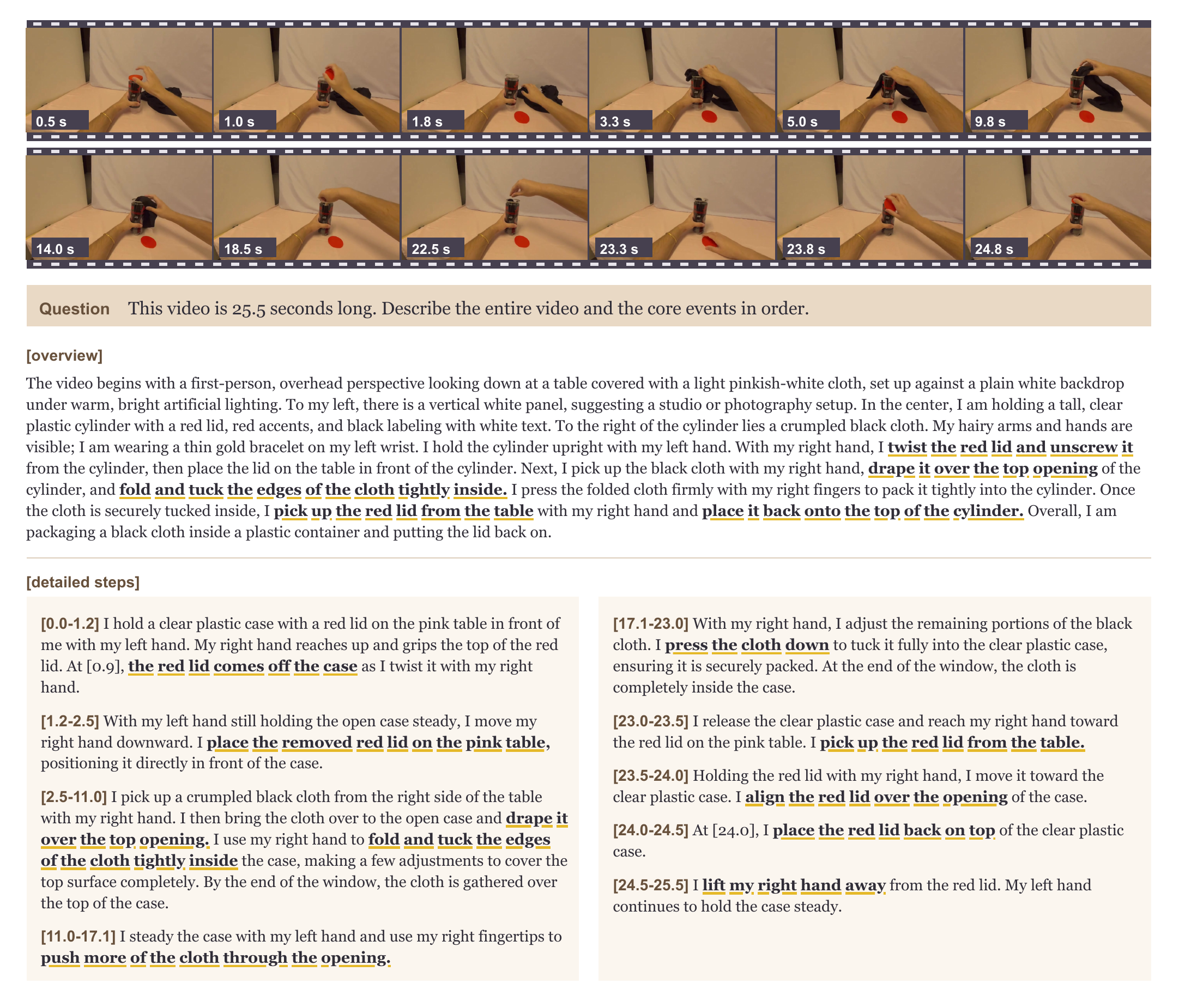}
\caption{Manipulation. The camera wearer removes the lid from a cylinder, packs a black cloth inside, and replaces the lid.}
\label{fig:supp_cases_types}
\end{figure}

\begin{figure}[!htp]
\centering
\includegraphics[width=\textwidth,height=0.85\textheight,keepaspectratio]{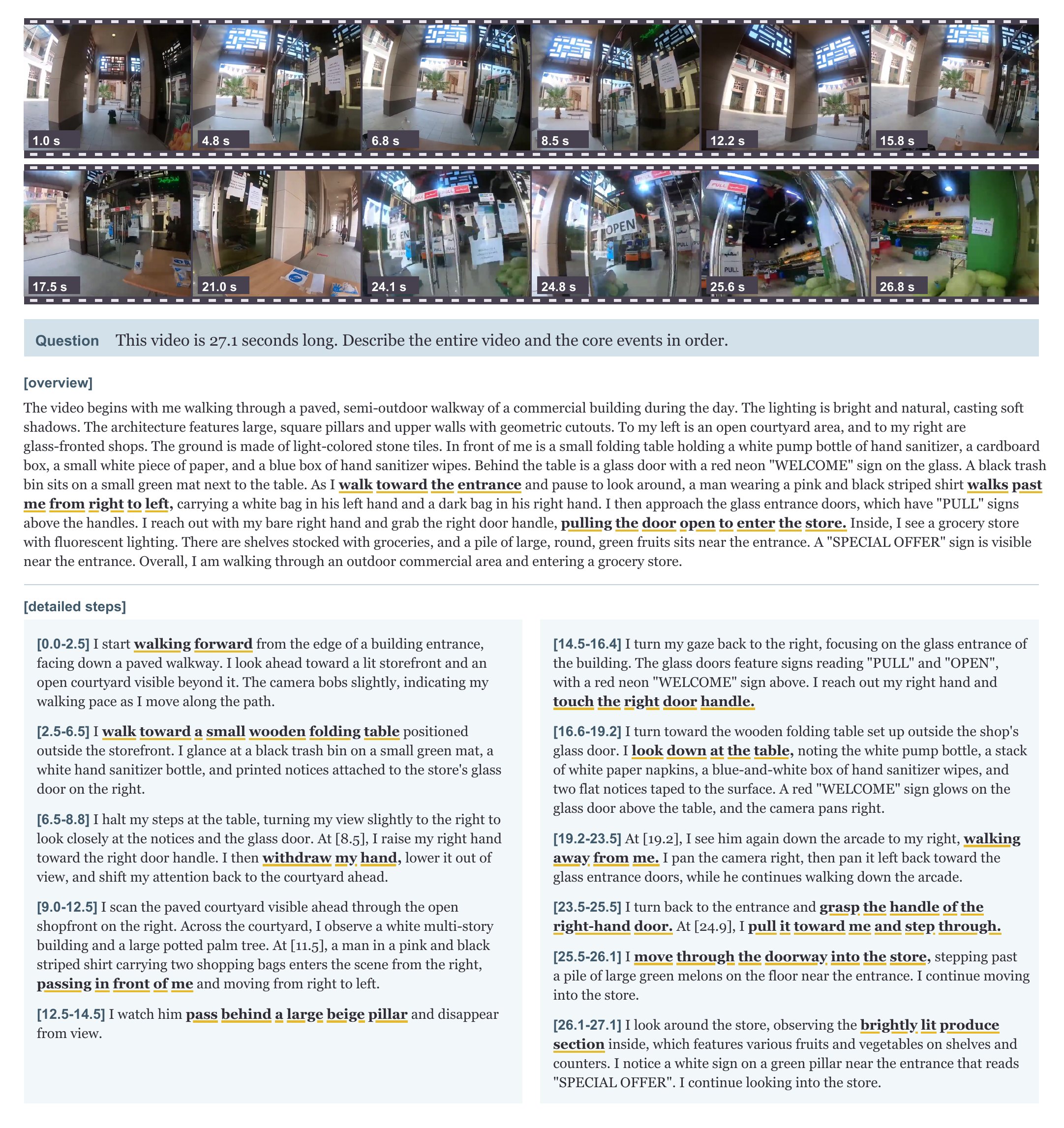}
\caption{Locomotion. The camera wearer follows a covered walkway, opens a glass door, and enters a grocery store.}
\label{fig:supp_case_locomote}
\end{figure}

\begin{figure}[!htp]
\centering
\includegraphics[width=\textwidth,height=0.85\textheight,keepaspectratio]{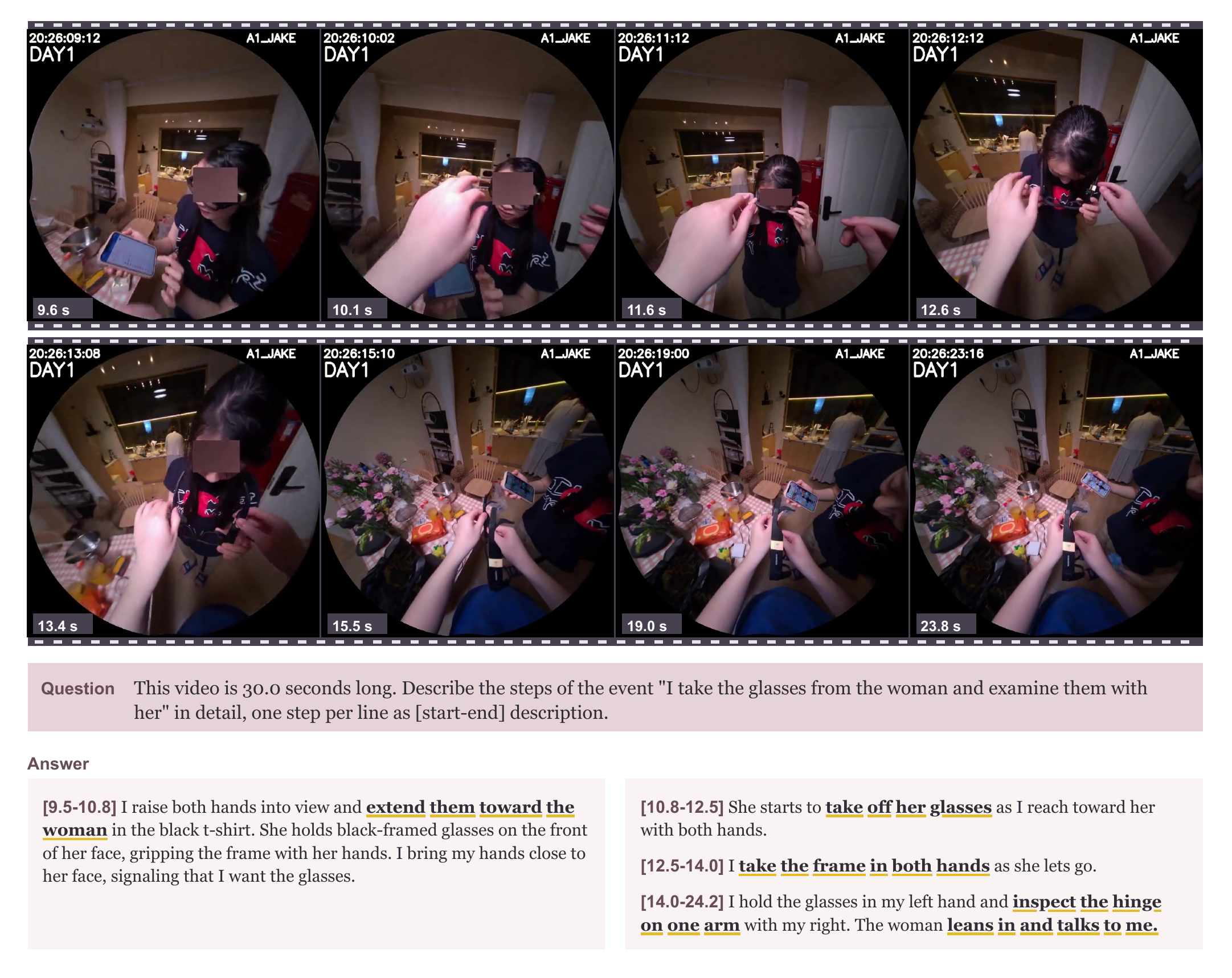}
\caption{Social interaction. The camera wearer takes a pair of glasses from a woman and examines the hinge while she leans in.}
\label{fig:supp_case_social}
\end{figure}

\begin{figure}[!htp]
\centering
\includegraphics[width=\textwidth,height=0.85\textheight,keepaspectratio]{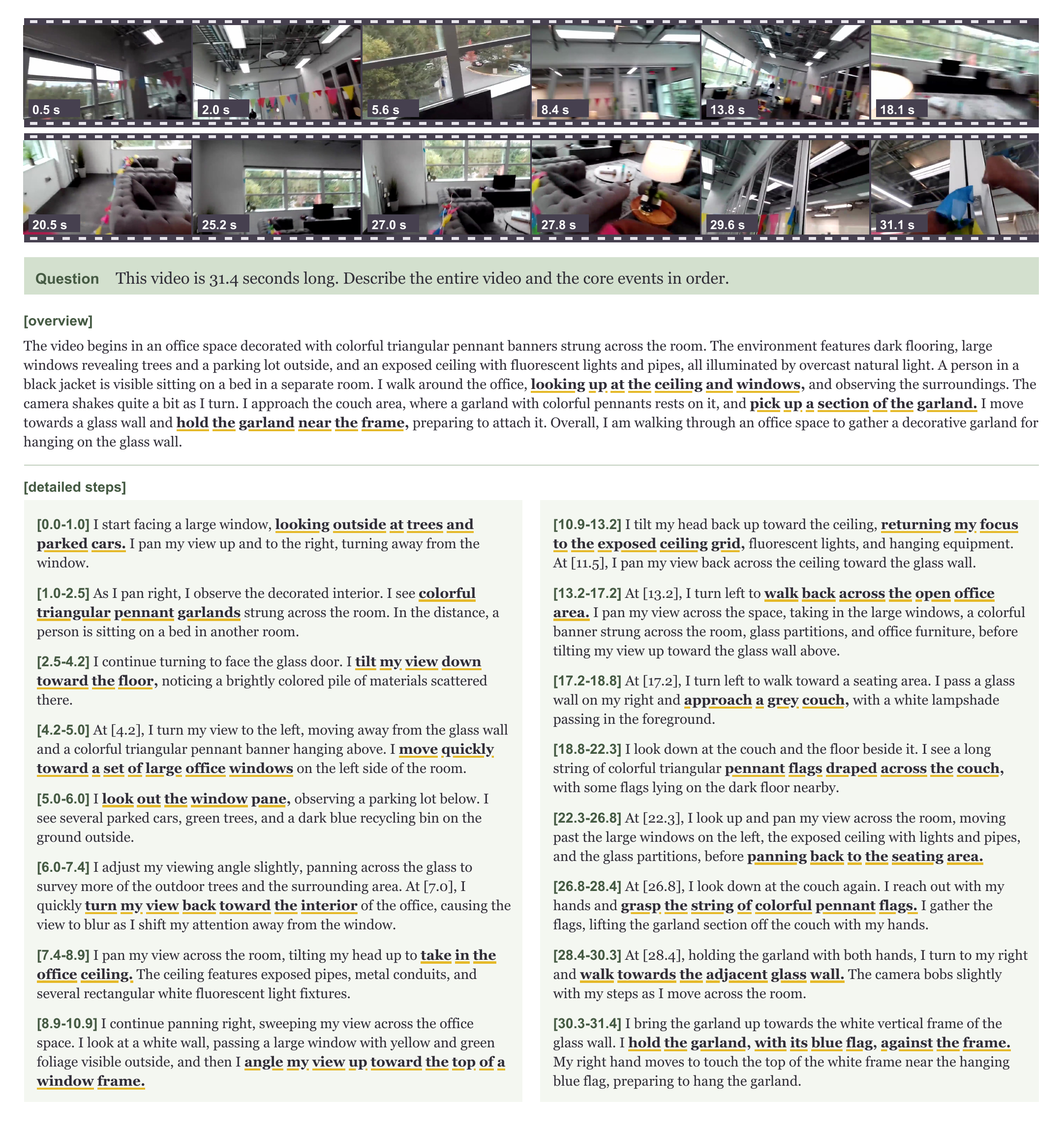}
\caption{Attention. The camera wearer looks around an office before picking up a pennant garland and holding it against a glass partition.}
\label{fig:supp_case_attend}
\end{figure}

Figures~\ref{fig:supp_cases_types}--\ref{fig:supp_case_attend} show four complete examples of \name captions. Each figure includes sampled video frames, the prompt, and the full model response. The manipulation, locomotion, and attention prompts ask for a description of the entire clip. The social prompt asks for the steps of a specific interaction.

\clearpage

\clearpage

\clearpage

\clearpage
\subsection{Coarse-to-Fine versus Single-Pass Training Annotations}
\label{sec:supp_cases_annot}
\begin{figure}[!htp]
\centering
\includegraphics[width=\textwidth,height=0.93\textheight,keepaspectratio]{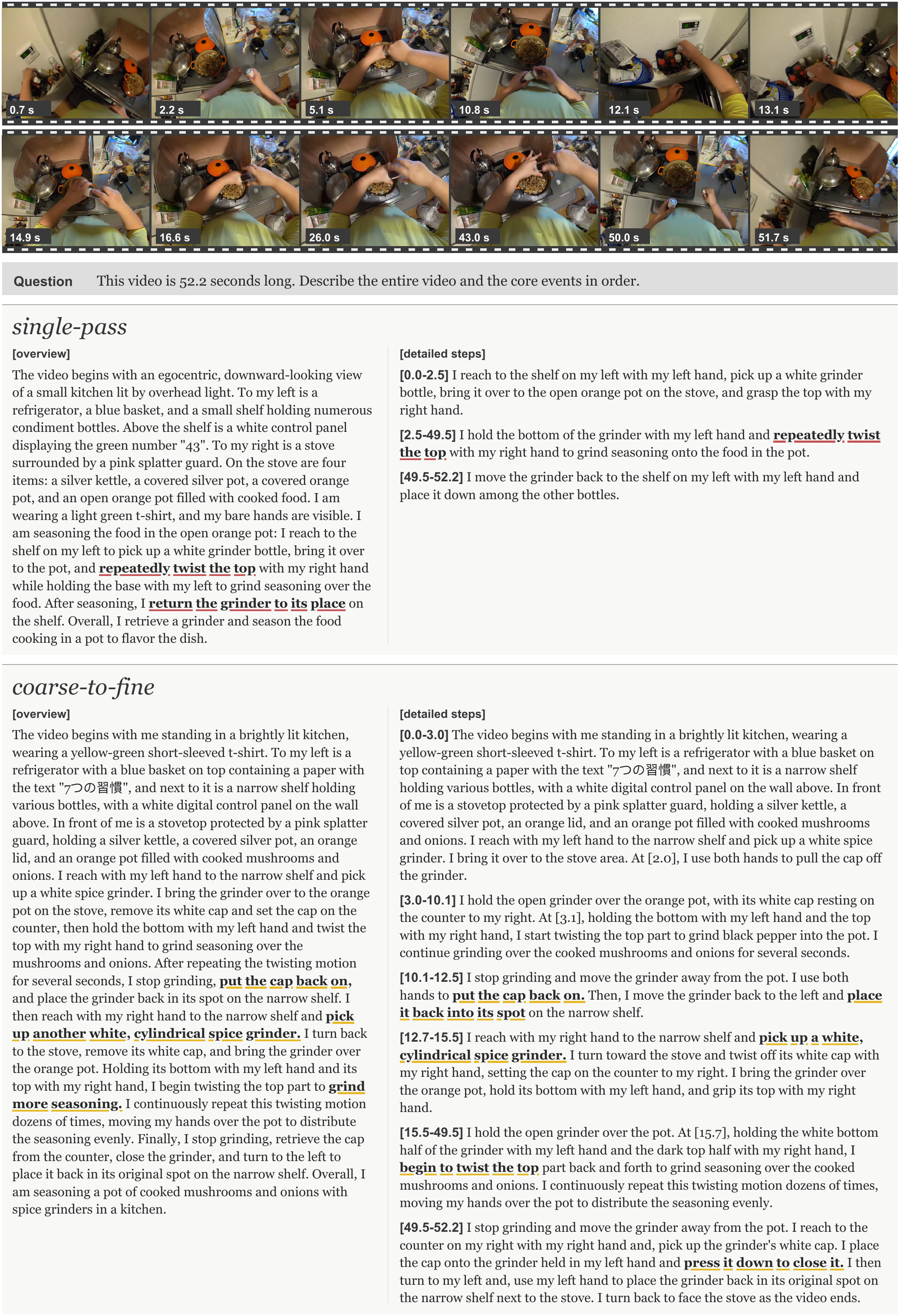}
\caption{Seasoning a pot. The single-pass annotation treats grinding as one continuous step. The coarse-to-fine annotation records when the wearer returns the first grinder and picks up the second.}
\label{fig:annot_grinders}
\end{figure}

We compare training captions generated by Gemini in a single pass with those produced by our coarse-to-fine pipeline for the same video and prompt. The examples cover an overview (Fig.~\ref{fig:supp_cases_annot}), an event-conditioned caption (Fig.~\ref{fig:annot_countertop}), and two full-video captions (Figs.~\ref{fig:annot_grinders} and~\ref{fig:annot_label}). Red and yellow underlines mark the phrases being compared in the single-pass and coarse-to-fine captions, respectively.
\par\smallskip
\noindent\begin{minipage}{\linewidth}
\centering
\captionsetup{type=figure}
\includegraphics[width=0.92\linewidth]{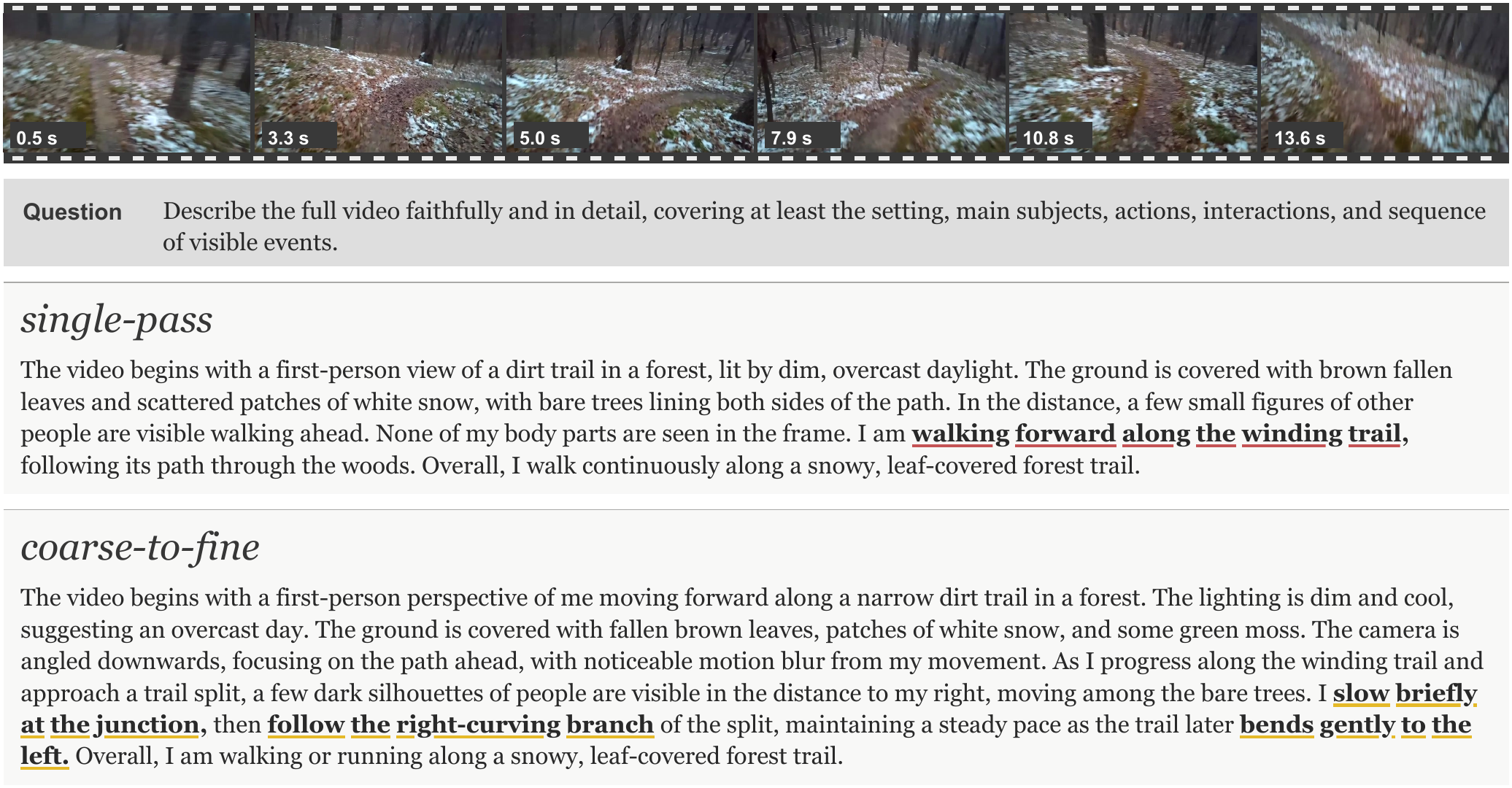}
\caption{Forest trail. Both overviews describe a winding path. The coarse-to-fine annotation identifies the right branch at the junction and the subsequent left bend.}
\label{fig:supp_cases_annot}
\end{minipage}\par
\medskip
\noindent\begin{minipage}{\linewidth}
\centering
\captionsetup{type=figure}
\includegraphics[width=0.92\linewidth]{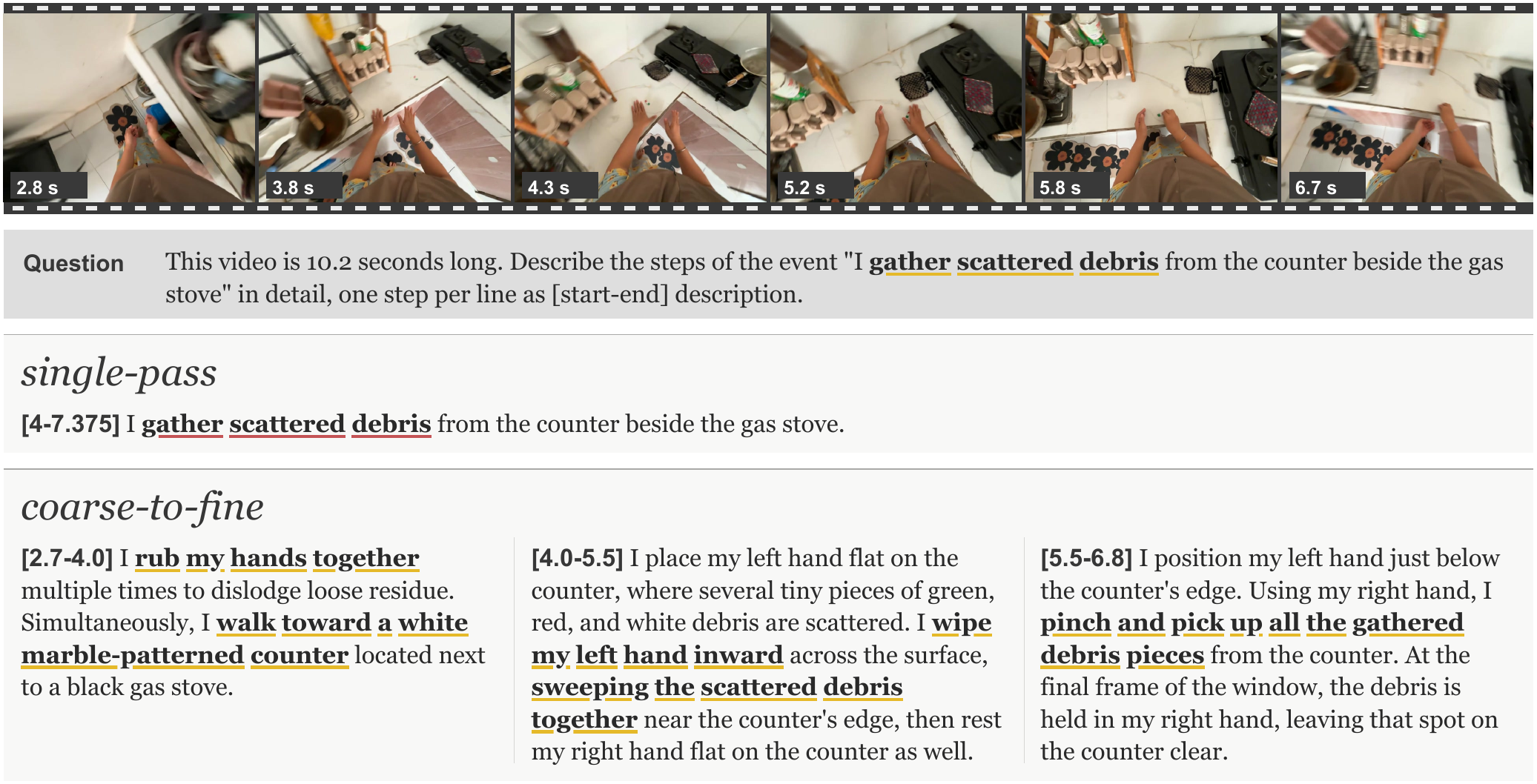}
\caption{Clearing a countertop. The single-pass caption summarizes the event in one interval. The coarse-to-fine caption separates the approach, the left-hand sweep, and the right-hand pickup.}
\label{fig:annot_countertop}
\end{minipage}\par

\clearpage
\clearpage
\noindent\begin{minipage}{\textwidth}
\centering
\captionsetup{type=figure}
\includegraphics[width=\textwidth]{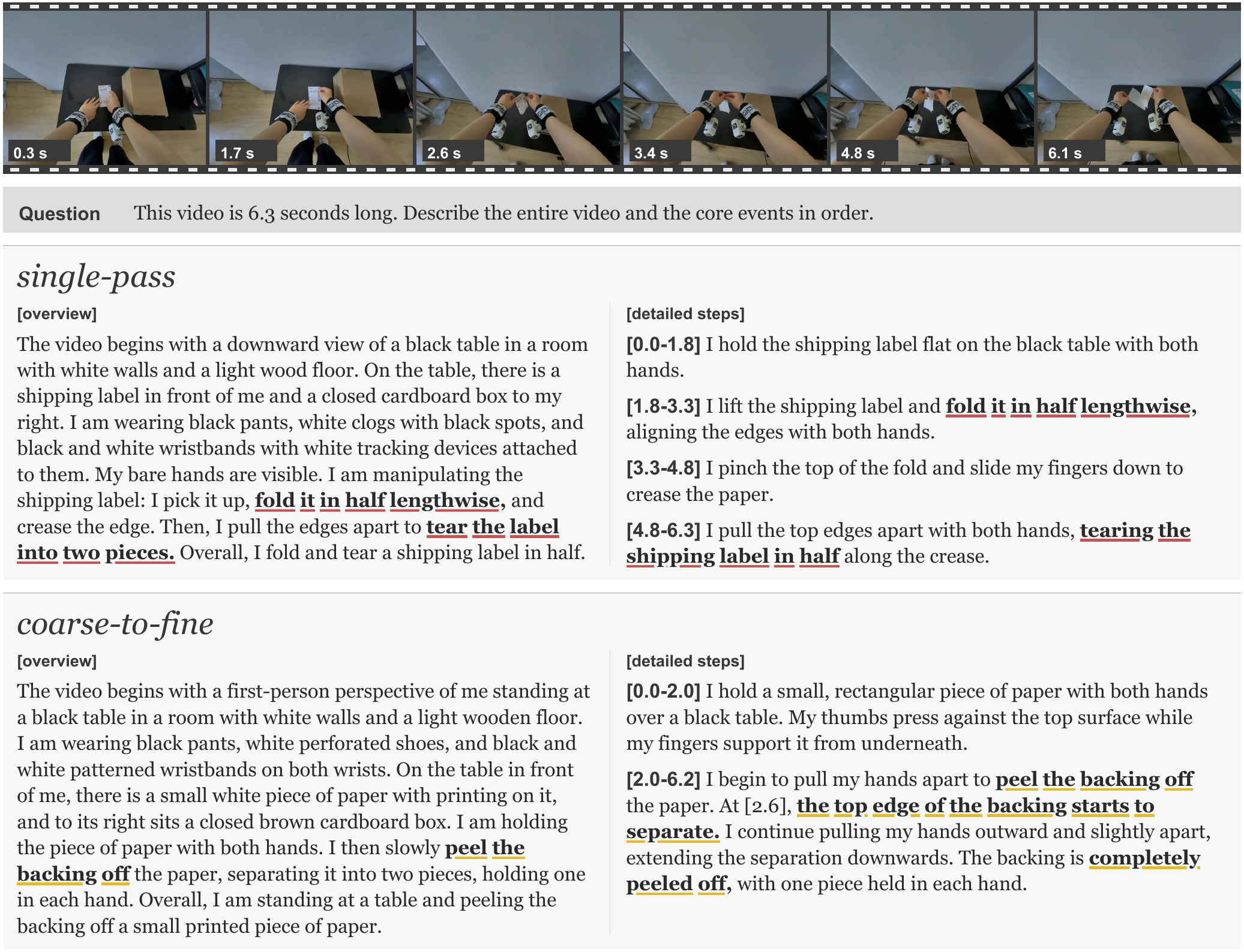}
\caption{Peeling a shipping label. The single-pass caption describes folding and tearing the label. The coarse-to-fine caption identifies the backing as the layer being peeled away.}
\label{fig:annot_label}
\end{minipage}\par

\section{Reward Details}
\label{sec:supp_reward}

\subsection{Fact Reward}
\label{sec:supp_reward_fact}
$R_{\mathrm{fact}}$ (Eq.~\ref{eq:fact_reward}) is computed in four steps. Facts are extracted from the text of each caption, pose-derived facts are added to the reference, the reference facts are deduplicated, and the facts of each caption are judged against the other caption and pooled into one F1.

\begin{table}[!ht]
\centering
\footnotesize
\caption{Fact families, the caption text each is extracted from, and their labels.}
\label{tab:supp_fact_families}
\setlength{\tabcolsep}{4pt}
\begin{tabularx}{\linewidth}{@{}l>{\raggedright\arraybackslash}p{0.28\linewidth}>{\raggedright\arraybackslash}X@{}}
\toprule
Family & Extracted from & Labels \\
\midrule
Motion & Steps, without At sentences & body: travel, posture, still. view: turn, roll, shake, still. \\
Focus & Steps, grouped by event, without At sentences & MANIP: operation, object. LOCOMOTE: travel mode, route. SOCIAL: partner, interaction. ATTEND: target, target state. \\
Macro & Overview & scene, main agent, role and purpose, activity arc, visual style, layout, legible text, people appearance, repetition count \\
Key moment & At sentences & -- \\
\bottomrule
\end{tabularx}
\end{table}

\paragraph{Extraction.}
Each family is extracted from the caption text listed in Table~\ref{tab:supp_fact_families}. Overview sentences give macro facts, which are extracted in two tiers, core and detail. The tier is used only during extraction and does not affect scoring. Step text gives motion and focus facts after every sentence containing an ``At [$t$]'' marker is removed, because those sentences are scored as key moments (Sec.~\ref{sec:supp_reward_km}). The program numbers the steps and the extractor cites only a step number, so each fact takes the time window written in the caption and the model never copies a time. Focus aspects depend on the event type, so steps are first grouped under the reference event they overlap most, and the reference and the prediction use the same grouping.

\paragraph{Pose facts as motion facts.}
When pose data are available, the pose-derived facts of Sec.~\ref{sec:supp_reward_pose} join the reference as motion facts. They carry the same body and view labels as facts extracted from text, so both kinds are judged and pooled in the same way. Body and camera movement is tracked geometrically by the pose annotations, whereas a caption written from video only estimates it. Pose facts are therefore the only reference facts that do not pass through caption text.

\paragraph{Deduplication.}
Mixing pose and caption facts creates two problems. Both can state the same movement, which would count it twice in recall, and they can disagree about it, which would reward matching the wrong one. A single language-model pass over all reference facts of a sample removes both (prompt in Sec.~\ref{sec:supp_prompt_judge}). The facts are listed in two groups, the steps group with motion and focus facts and the overview group with macro facts. Each fact is tagged as caption or sensor, where sensor marks pose-derived facts, and carries its time range. A fact is removed in two cases.
\begin{itemize}
    \item \emph{Duplicate.} Two facts in the same group state the same thing about the same moment. The less specific one is removed, and a pose fact is kept over a caption fact that says no more.
    \item \emph{Conflict.} Two facts, in the same group or across groups, cannot both be true of the same person, object or movement at the same time. The one less likely to be correct is removed, with a preference for the pose fact on camera and body movement. Both are removed when the pass cannot tell.
\end{itemize}
Facts at different times, facts about different objects or people, and facts that only add or omit details of each other are kept. The model only names the facts to remove and does not rewrite any, and the pass runs once when the reference is built.

\paragraph{Judging.}
Reference facts are judged against the predicted caption, and predicted facts against the reference caption. Each request covers one caption section. Motion and focus facts are judged together against the steps as written, with the At sentences left in place so that the caption reads coherently, and macro facts against the overview alone, so a macro fact cannot earn credit from a step. Facts are listed under their time ranges and matched to the caption by content, since timestamps can differ slightly. The judge reads each fact by its sentence structure. The subject, complement, verb and object that the sentence has form its core, and a core part matches when the caption uses the same word, a synonym, or a pronoun that clearly refers back to it (prompt in Sec.~\ref{sec:supp_prompt_judge}). The verdict maps to credit $s(f\mid y)$ as follows.
\begin{itemize}
    \item \emph{Supported} (1). Every core part and detail matches, whatever the wording.
    \item \emph{Partial} (0.4). Every core part matches, but the caption leaves out some details.
    \item \emph{Contradicted} (0). The fact and the caption cannot both be true under any reading.
    \item \emph{Absent} (0). None of the above.
\end{itemize}

\paragraph{Pooling.}
Facts of all families and the key moments are pooled into one F1 instead of one F1 per family. Then
\begin{equation}
    p=\frac{\sum_{f\in\mathcal F(\hat y)}s(f\mid y^\star)}{|\mathcal F(\hat y)|},\qquad
    r=\frac{\sum_{f\in\mathcal F(y^\star)}s(f\mid\hat y)}{|\mathcal F(y^\star)|},
    \label{eq:supp_fact_pr}
\end{equation}
and $R_{\mathrm{fact}}$ is their F1 as in Eq.~\ref{eq:fact_reward}. Key moments enter these sums as described in Sec.~\ref{sec:supp_reward_km}.
\subsection{Pose-Derived Motion Facts}
\label{sec:supp_reward_pose}
Pose-derived facts come from two kinds of recordings. EgoPro and HumanPlus provide a body skeleton and a head orientation for every frame, while Sekai provides only a camera trajectory estimated from the video.
We turn these measurements into text in three steps.
Each frame value is first assigned to a discrete bin.
A stretch of frames over which the bin changes in one direction then becomes a motion event, and the bin difference between its first and last frame gives the event's category.
Finally, each selected event is written as a short sentence together with its measured time span.
The bins, event categories and candidate lists follow MotionScript~\cite{yazdian2025motionscript}. The camera bins and the few extra thresholds stated below are our own.

\paragraph{From frames to bins.}
For the skeleton, we measure the pelvis translation along the left--right, up--down and forward--backward axes of the body's horizontal heading.
We also measure the torso's forward lean (pitch), its sideways tilt (roll) and its turn (yaw) relative to its orientation at the start of the clip.
The head is described by the forward axis of the head joint expressed in the torso frame, which gives its yaw and pitch relative to the body.
Measuring the head against the torso keeps body turns and walking out of the head motion.
Each track is smoothed with a median filter over at most seven frames and binned with the MotionScript thresholds.
Translation uses 5\,cm bins. Lean, tilt and turn use the MotionScript angle bins, for example $\pm25^\circ$, $\pm60^\circ$, $\pm90^\circ$ and $\pm135^\circ$ for the turn.
Invalid frames and temporal gaps split a track into separate pieces.

\paragraph{From bins to events.}
Within each piece we follow the bin index over time.
An event starts once the bin leaves its starting value and ends at the furthest bin reached before the track turns back by more than one bin.
An event is kept only when its total change exceeds one bin.
The signed bin difference between its first and last frame gives the direction and the magnitude, which the MotionScript motioncode thresholds grade from very short to very long for translations and from slight to significant for rotations.
A very long body translation, for instance, spans more than ten bins, that is, more than 0.5\,m.

\paragraph{Camera trajectories.}
Sekai clips have no skeleton, so we use the relative pose between consecutive camera frames.
Each relative pose is expressed in the axes of the camera at the end of that step, so left, right, up and down always follow the current view.
Per-axis translation and rotation velocities are averaged over a 0.2\,s window, and values below a noise floor are set to zero.
A pause longer than 0.5\,s or a reversal larger than one bin ends an event.
We drop events shorter than 0.15\,s and events whose net change is less than 65\% of their total path.
Camera translation has no metric scale, so one translation bin is 0.2 of the clip's median distance traveled per second of motion.
One rotation bin is $15^\circ$, and any rotation of at least $20^\circ$ is graded as at least moderate.

\paragraph{Selection and wording.}
Table~\ref{tab:supp_pose_facts} lists the events that become facts and the sentence written for each.
Head facts also require an angle change of at least $35^\circ$, because smaller head turns are hard to see in the video.
A head-mounted camera cannot tell a torso turn from a head turn, so a torso turn is written as the view turning, and it is dropped when a head turn in the same direction overlaps it.
When a skeleton clip contains no body translation or torso turn fact, we add a single fact for the whole clip stating that I stay in the same place.
Each fact carries its measured span, rounded to 0.01\,s, and the type, sub-type and direction labels used for caption motion facts (Sec.~\ref{sec:supp_prompt_judge}). Lowering or raising the body is labeled as posture rather than travel.

For an event-conditioned example, we keep a pose fact when at least half of its span lies inside the event.
The pose facts and the reference caption facts then go through a single deduplication pass (Sec.~\ref{sec:supp_prompt_judge}), which removes repeated or conflicting facts and prefers the pose fact for camera and body movement.
The judge sees each remaining pose fact under its time range, like any other reference fact.

\begin{table}[!ht]
\centering
\footnotesize
\caption{Pose-derived motion facts. Categories follow MotionScript.}
\label{tab:supp_pose_facts}
\setlength{\tabcolsep}{4pt}
\begin{tabularx}{\linewidth}{@{}l>{\raggedright\arraybackslash}p{0.24\linewidth}>{\raggedright\arraybackslash}p{0.20\linewidth}>{\raggedright\arraybackslash}X@{}}
\toprule
Source & Measurement & Kept categories & Sentence \\
\midrule
Skeleton & Body translation, horizontal & very long ($>0.5$\,m) & I move to the left / right. I move forward / backward. \\
 & Body translation, vertical & very long ($>0.5$\,m) & I move downward / upward. \\
 & Torso turn & slight, moderate, significant & The view turns left / right. \\
 & Torso lean & slight, moderate, significant & I bend / lean my upper body forward. I straighten / lean my upper body back. \\
 & Torso tilt & moderate, significant & I tilt my body to the left / right. \\
 & Head turn ($\geq35^\circ$) & slight, moderate, significant & I turn my head to the left / right. \\
 & Head pitch ($\geq35^\circ$) & slight, moderate, significant & I look down / up. \\
 & No translation or torso turn & whole clip & I stay in the same place. \\
\midrule
Camera & Rotation & $\geq20^\circ$ & The camera turns left / right, tilts upward / downward, or rolls clockwise / counterclockwise. \\
 & Translation, horizontal & long, very long & The camera moves forward / backward, or moves to its left / right. \\
\bottomrule
\end{tabularx}
\end{table}

\subsection{Key-Moment Scoring}
\label{sec:supp_reward_km}
Each ``At [$t$],'' sentence in a predicted step is one predicted key moment, and the key moments of the reference caption form the reference set.
Each predicted moment is classified as correct or wrong in the following order.

\paragraph{Placement.}
A moment whose time lies outside the window of its own step is wrong, and no judge is called.

\paragraph{Reference match.}
The reference moments within 1\,s of $t$ are its candidates.
A text judge (Sec.~\ref{sec:supp_prompt_judge}) decides for each candidate whether the predicted sentence describes the same moment, and a confirmed moment is correct.
Confirmed pairs are assigned one to one, nearest in time first, so each reference moment credits a single prediction.
A confirmed prediction whose reference went to a nearer prediction scores zero and is not counted as an error.

\paragraph{Unmatched moments.}
When no reference confirms a moment, a second text judge decides whether the sentence states one of the key-moment types defined for annotation (Sec.~\ref{sec:supp_prompt_annot}).
A sentence that does not is wrong.
A sentence that does is checked against frames sampled at 10\,fps from $[t-2,\,t+2]$\,s and clipped to its step.
The frame judge decides whether the stated event happens and marks the frames where it happens.
The moment is wrong when the event does not happen, or when the nearest marked frames are more than 1\,s from $t$.
Otherwise it is a correct key moment that the reference missed, and it earns the same credit as a reference match.

\paragraph{Credit.}
A correct moment earns $e^{-d/\tau}$ with $\tau=2$\,s, where $d$ is the distance from $t$ to the matched reference time, or to the nearest marked frame span for a moment verified from frames.
With $C^{+}$ the summed credit of correct moments and $E$ the number of wrong moments, the key-moment credit is
\begin{equation}
    C_{\mathrm{km}}=\max\bigl(0,\,C^{+}-E\bigr),
    \label{eq:supp_km_credit}
\end{equation}
so each wrong moment cancels one exactly timed correct moment.

\paragraph{Pooling with facts.}
Key moments enter $R_{\mathrm{fact}}$ (Eq.~\ref{eq:fact_reward}) as one more fact family.
With $n$ predicted and $m$ reference moments, this family adds $C_{\mathrm{km}}$ to the precision numerator and $n$ to its denominator, and adds $\min(C_{\mathrm{km}},m)$ to the recall numerator and $m$ to its denominator.
When the reference has no key moment, key moments enter precision only.

\subsection{Temporal Reward, Format Reward and Validity Check}
\label{sec:supp_reward_time}
\paragraph{Validity check.}
Before any judge is called, a parser reads the rollout in the output format of its task.
The rollout receives $R=-1$ when any of the following holds.
\begin{itemize}
    \item Generation stops at the limit of 16{,}384 tokens.
    \item The text contains a verbatim loop, a block of 4 to 200 words repeated three times in a row. Case and text in square brackets, such as timestamps and section labels, are ignored in this check.
    \item A full-video caption does not contain exactly one overview section followed by one step section, or either section is empty. An overview caption is empty.
    \item No step is found, a step line has no $[\mathrm{start}\text{--}\mathrm{end}]$ window, or a window has no description.
    \item A step starts before 0\,s or at or after the video duration, ends more than 0.05\,s after the duration, or does not start before it ends. An end within 0.05\,s after the duration is treated as rounding and clipped.
    \item Steps are not written in temporal order, or two step windows overlap. Adjacent windows and gaps between them are allowed.
\end{itemize}
Key-moment times are not part of this check. A key moment outside its step is scored as wrong (Sec.~\ref{sec:supp_reward_km}).

\paragraph{Temporal reward.}
$R_{\mathrm{time}}$ (Eq.~\ref{eq:time_reward}) uses only the step windows.
The tIoU of two intervals is the length of their intersection divided by the length of their union.
All predicted--reference pairs are sorted by tIoU in descending order, and a pair is accepted when neither of its intervals has been used. Matching stops at the first pair with zero overlap.
The sum is divided by the larger of the two step counts, so steps left over on the longer side add to the denominator without adding to the sum, and splitting a reference step or merging several reference steps both lower the reward.

\paragraph{Format reward.}
Overview captioning has no step windows, so $R_{\mathrm{format}}$ takes the place of the temporal and style rewards. It averages two checks on the overview text, whether it opens with ``The video'' and whether its last sentence begins with ``Overall,'', and each check scores 1 when it holds and 0 otherwise. An overview that contains a step window or an ``At [$t$]'' marker receives $R_{\mathrm{format}}=0$, since an overview is plain scene-level text.

\subsection{Optimization Settings}
\label{sec:supp_reward_opt}
SFT trains LoRA adapters on all linear layers of the language model and keeps the vision encoder and the vision--language projector frozen.
Its learning rate warms up linearly over the first 5\% of steps and then decays linearly.
GRPO runs one epoch over the 8{,}000 training pairs, which gives 250 steps. Each step takes 32 prompts with 8 rollouts each and makes a single gradient update on all 256 responses.
The learning rate warms up linearly over the first 10 steps and then stays constant.
Rollouts are sampled with temperature 1.0, top-$p$ 0.95 and top-$k$ 20.
The KL term uses the low-variance estimator.
We collect rollouts asynchronously, pausing unfinished generations before each actor update and resuming them with the updated weights. The log-probabilities of each token under the policy that generated it are stored with the response. A response is used only if the earliest policy version that generated any of its tokens is at most two updates older than the policy being updated, and a complete group that exceeds this limit is dropped. The $G$ responses of a group are still normalized together. Before each update, the trainer recomputes token log-probabilities under the pre-update policy $\pi_{\theta_{\mathrm{old}}}$, which supplies the denominator of $\rho_{i,t}$ in Eq.~\ref{eq:grpo_objective}. We use dual-clip PPO with $c=3$ for negative advantages. After clipping, we multiply each token's policy-gradient term by $w_{i,t}=\min\bigl(\pi_{\theta_{\mathrm{old}}}(\hat y_{i,t}\mid x,\hat y_{i,<t})/\pi_{\mathrm{gen}}(\hat y_{i,t}\mid x,\hat y_{i,<t}),\,2\bigr)$, where $\pi_{\mathrm{gen}}$ is the policy that generated that token. We treat $w_{i,t}$ as constant during differentiation.
Both stages sample video at a target rate of 4\,fps, capped at 240 frames.
\begin{table}[!ht]
\centering
\footnotesize
\caption{Training hyperparameters.}
\label{tab:supp_hparams}
\begin{tabular*}{\columnwidth}{@{\extracolsep{\fill}}lcc@{}}
\toprule
 & SFT & GRPO \\
\midrule
Trainable parameters & \multicolumn{2}{c}{LoRA, rank 32} \\
Learning rate & $1{\times}10^{-4}$ & $1{\times}10^{-5}$ \\
Epochs / steps & 2 epochs & 250 steps \\
Batch size & 16 & 32 \\
Rollouts per input $G$ & -- & 8 \\
KL coefficient $\beta$ & -- & 0.02 \\
Clip range $\varepsilon$ & -- & 0.2 \\
Max frames / fps & \multicolumn{2}{c}{240 / 4} \\
GPUs & 16 & 16 \\
\bottomrule
\end{tabular*}
\end{table}

\section{Training Data Statistics}
\label{sec:supp_data}
We deduplicated the training data against every evaluation set by video ID and time segment, so no training clip shares a source video segment with an evaluation item.
For HumanPlus-1000, we used the public preview release~\cite{humanplus1000}. We took the left-camera stream from each stereo recording and split it into clips of at most 30\,s. This yielded 236 clips from 88 recordings, with a mean duration of 24.44\,s.

\begin{table}[!ht]
\centering
\footnotesize
\caption{Composition of the SFT mix.}
\label{tab:supp_data_sources}
\setlength{\tabcolsep}{3pt}
\begin{tabular*}{\columnwidth}{@{\extracolsep{\fill}}lrrr@{}}
\toprule
Source & Videos & Mean duration (s) & Examples \\
\midrule
Ego4D Goal-Step & 2{,}354 & 52.77 & 6{,}019 \\
EgoHOD (Ego4D) & 5{,}166 & 15.80 & 9{,}909 \\
Sekai & 2{,}743 & 47.61 & 6{,}860 \\
EgoVid-5M & 1{,}337 & 4.07 & 2{,}670 \\
EgoSuite (EgoPro) & 3{,}876 & 27.46 & 10{,}136 \\
HumanPlus-1000 & 236 & 24.44 & 472 \\
RekaDaily-10k & 4{,}503 & 13.17 & 10{,}933 \\
EgoLife & 1{,}538 & 29.83 & 3{,}001 \\
\midrule
Total & 21{,}753 & 25.71 & 50{,}000 \\
\bottomrule
\end{tabular*}
\end{table}

\begin{table}[!ht]
\centering
\footnotesize
\caption{Reference captions by event type across 77{,}382 events. An event can have multiple labels, so the shares can sum to more than 100\%.}
\label{tab:supp_data_types}
\setlength{\tabcolsep}{2pt}
\begin{tabular*}{\columnwidth}{@{\extracolsep{\fill}}lrrrr@{}}
\toprule
Event type & Share (\%) & \shortstack[r]{Steps /\\event} & \shortstack[r]{Words /\\step} & \shortstack[r]{Key moments /\\step} \\
\midrule
\textsc{manip} & 70.33 & 3.51 & 43.43 & 0.70 \\
\textsc{locomote} & 31.50 & 3.46 & 51.37 & 0.96 \\
\textsc{social} & 6.88 & 4.41 & 67.75 & 1.46 \\
\textsc{attend} & 32.01 & 3.56 & 48.28 & 0.90 \\
\bottomrule
\end{tabular*}
\end{table}

\begin{figure}[!ht]
\centering
\includegraphics[width=0.62\linewidth]{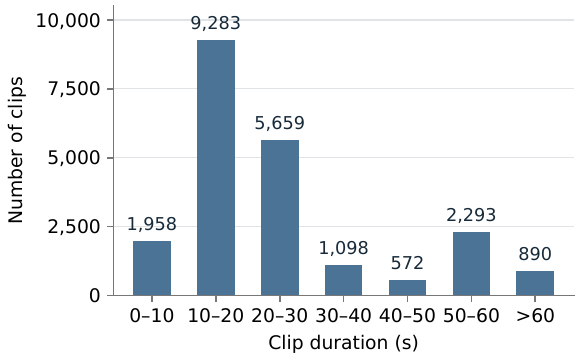}
\caption{Duration of the 21{,}753 unique video clips. Bars show clip counts in 10-second intervals up to 60\,s and a single bin for longer clips. Each interval includes its upper endpoint.}
\label{fig:supp_clip_durations}
\end{figure}
\FloatBarrier

\section{Evaluation Setup}
\label{sec:supp_eval}

\paragraph{Decoding.}
\name generates every caption with temperature 1.0, top-$p$ 0.95 and top-$k$ 20, with thinking disabled and at most 16{,}384 new tokens. Video is sampled at 4\,fps with at most 240 frames, as in training.

\paragraph{Instructions.}
The instruction templates are listed in Sec.~\ref{sec:supp_prompt_task}.

\section{Prompts}
\label{sec:supp_prompts}
This section collects the core instructions of the prompts used in our pipeline. Sec.~\ref{sec:supp_prompt_task} lists the instructions given to \name for the three captioning tasks. Sec.~\ref{sec:supp_prompt_annot} lists the annotation prompts in the order of Fig.~\ref{fig:data_construction}(a)(b), followed by the prompt of the single-pass baseline in Sec.~\ref{sec:ablation}. Sec.~\ref{sec:supp_prompt_qa} lists the prompts of the caption-to-QA protocol in Sec.~\ref{sec:exp_setup}. Sec.~\ref{sec:supp_prompt_judge} lists the prompts used to compute the rewards of Sec.~\ref{sec:supp_reward}, namely fact extraction, judging and deduplication, key-moment judging, and the two style judges. Text in braces, such as \texttt{\{video\_duration\_s\}}, is filled in by the program for each sample, and the tags in angle brackets only separate the parts of a prompt.

\subsection{Task Instructions}
\label{sec:supp_prompt_task}
\name receives one short instruction for each captioning task of Sec.~\ref{sec:task_form}, and these templates are used in training. Text in braces is filled in by the program, and the video duration is rounded down to one decimal place. The event-conditioned template names the event without a time window. At evaluation, EgoMe uses its own instructions, one for Overview and one for Steps, which are given in the two EgoMe boxes below, and \name uses the templates above on the other benchmarks. The comparison models receive the same instruction as \name with two exceptions. On EgoHome, \name receives the full-video template, whereas the proprietary models and the external baselines receive the EgoMe Steps instruction, because the full-video template does not ask for timestamps and PSJ can only score timed steps. On PhysCapBench, \name receives the full-video template, whereas the 14 comparison models receive the physical-event instruction in the last box below.

\begin{promptbox}{Full-video captioning}
This video is {duration} seconds long. Describe the entire video and the core events in order.
\end{promptbox}

\begin{promptbox}{Overview captioning}
Describe the full video faithfully and in detail, covering at least the setting, main subjects, actions, interactions, and sequence of visible events.
\end{promptbox}

\begin{promptbox}{Event-conditioned captioning}
This video is {duration} seconds long. Describe the steps of the event "{event}" in detail, one step per line as [start-end] description.
\end{promptbox}

\begin{promptbox}{EgoMe Overview}
Describe what the camera wearer does throughout the video in English. Summarize the main actions in order, the objects involved, and their visible changes. Include object details and hand use when relevant. Describe the actions as they occur. Return only the paragraph.
\end{promptbox}

\begin{promptbox}{EgoMe Steps}
This video is {duration} seconds long. Describe my action steps in order, including repeated ones, one step per line as [start-end] description.
\end{promptbox}

\begin{promptbox}{PhysCapBench}
Describe the visible physical events in chronological order. Include objects, actions, interactions and visible changes. Do not invent hidden causes.
\end{promptbox}

\subsection{Coarse-to-Fine Annotation}
\label{sec:supp_prompt_annot}

\begin{promptbox}{Overview}
<Task>
Describe the entire video faithfully and in detail, including the setting, main subjects, their actions and interactions, and how the visible events unfold.
</Task>

<Input>
The attached video is {video_duration_s} seconds long.
</Input>

<Rules>
Describe the full input clip in natural English, using “I” for the camera wearer.
Structure:
- Open with “The video begins/opens/starts with...” to describe the opening scene.
- Close with one sentence beginning “Overall, ...” to summarize my overall activity and its purpose.

Content requirements:

1. Establish the world context first: name the place and scene type, the surfaces,
   equipment, and lighting around me, and where the main objects and people sit
   relative to me. Describe this before the activity, weaving the environmental details below into the description:

- **Look and light** — the time of day and lighting (source, direction, brightness, warm or cool cast); the dominant colors and color palette or grading; noticeable image quality (motion blur, shallow depth of field) or a stylized look (time-lapse, slow motion); and how the camera is used — perspective and shot type (overhead, close-up, wide), movement, and shifts in what it faces, such as a split screen.
- **Setting and layout** — the place and scene type, its concrete surroundings (backgrounds, ground or floor materials, road markings, buildings, equipment), where key items sit relative to me (left/right, in front, behind) and to each other, and the exact count of salient objects (e.g., three bowls on the rack).
- **Text** — reliably legible text on signs, labels, screens, and brand names, copied EXACTLY in its original language; report clock readings when visible.
- **People** — the visible appearance of the camera wearer (clothing, sleeves, hands, footwear, a visible reflection) and the main people with their activities and visible emotional expressions, each identified by one consistent description.
  Cover every category above that the video visibly supports.

2. Describe events chronologically at the activity level: state the main subjects' activity and its visually supported purpose before describing supporting actions, interactions, and changes in objects or the scene.
3. For repeated actions, state the number of visible occurrences before describing them. Group similar repetitions together while preserving chronology.

<Format>
Return JSON only:
{"dense_caption": "Full-video description without timestamps."}
Always include dense_caption.
</Format>
\end{promptbox}

\begin{promptbox}{Event structure and step outline}
<Task>
Identify up to {max_events} distinct events from the attached video and outline the sub-steps of each.
</Task>

<Input>
- Video duration: {video_duration_s} seconds.
</Input>

<EventTypes>
Classify each event with one or more of these labels:
- MANIP: manipulation of an object by hand or with a tool.
- LOCOMOTE: movement through space, including passive movement such as riding.
- SOCIAL: communication or coordinated action between people.
- ATTEND: viewing, watching, or visually following people, objects, or scenes.
</EventTypes>

<OutputFormat>
Return JSON only with the exact fields shown. Times are in seconds from video start, one decimal place, sorted by start time.
- event: a short, natural sentence naming the actor and action, such as "I sort the plates" or "The man carries a box". Use "I" for the camera wearer and a consistent visual description for another person. Base the actor and action on the video.
- steps: the event's distinct sub-tasks in chronological order. Each step should be a meaningful action that can be further broken down — finer than the event itself, but coarser than atomic motions like "raising a hand".

IMPORTANT: Events must not overlap in time. Each event's end time must be less than or equal to the next event's start time. Arrange events sequentially with no temporal overlap.

{
  "candidates": [
    {
      "event": "I open the laptop on the table",
      "event_type": ["MANIP"],
      "start": "2.0",
      "end": "5.4",
      "steps": [
        "positioning both hands on the laptop",
        "lifting the laptop lid open",
        "releasing the lid and withdrawing my hands"
      ]
    }
  ]
}
</OutputFormat>
\end{promptbox}

\begin{promptbox}{Step description}
<Task>
Describe the event in <Events> inside its assigned window. Expand each outlined step into one timestamped paragraph, and describe visible actions the outline misses inside the paragraph where they happen. If the outline disagrees with the video, the video wins.
</Task>

<Input>
- Video duration: {duration} seconds.
- The clip is this event plus up to one second of context on each side. All times are crop-local seconds.
- Describe only what happens inside the assigned window.
</Input>

<Events>
<EventTemplate>
<Event candidate_id="{candidate_id}">
event: {event}
event_type: {event_type_json}
Assigned window: [{event_start}-{event_end}]
Step outline: {steps_json}
</Event>
</EventTemplate>
</Events>

<DescriptionFocus>
Cover the aspects below that match the event's labels. When one moment fits several aspects, describe it once.

<When event_type="MANIP">
MANIP — object manipulation
| Aspect         | What to describe                                                       |
|---|---|
| Actions        | What I do to which object, tools used, sequence of operations            |
| Hand/body      | Each hand's role, grip, contact, direction of movement                  |
| Object changes | How objects change in position, shape, contents, or attachment          |
| Placement      | Where objects are taken from and put down — side, shelf/row, container, relative to nearby objects |
| End state      | Object state at the final frame of the assigned window                  |
| Errors         | Adjustments, retries, slips — include when visible                      |
</When>

<When event_type="LOCOMOTE">
LOCOMOTE — movement through space
| Aspect         | What to describe                                                       |
|---|---|
| Route          | Start location, areas passed, turns, exits, last visible location       |
| Travel mode    | How I travel — walking, running, cycling, riding — and any pace change, stop, or restart. Maybe use cues such as a wheel in the view/an upright or seated posture...|
| Camera motion  | Camera bobbing, sway, pans and tilts. Say which of it comes from my body and what it suggests about the mode, and which is only the camera turning; never read a direction change out of camera motion alone |
| Spatial cues   | Landmarks, signs, readable text that identify locations, placed relative to the path |
</When>

<When event_type="SOCIAL">
SOCIAL — interpersonal interaction
| Aspect         | What to describe                                                       |
|---|---|
| Who and what   | Participants, what the interaction concerns                             |
| Sequence       | What each person does toward others, in order; turn-taking, handoffs    |
| Responses      | How others react, how the interaction ends or continues                  |
| Signals        | Gestures, expressions, distance changes — include when they clarify intent |
</When>

<When event_type="ATTEND">
ATTEND — viewing and observation
| Aspect         | What to describe                                                       |
|---|---|
| Targets        | What is being viewed, its location and distinguishing features          |
| Content        | What targets are doing or showing, how they change                      |
| View process   | Attention switches, obstructions, targets leaving/reappearing          |
</When>
</DescriptionFocus>

<CommonRules>
- One [start-end] paragraph per outlined step, in order, separated by blank lines.
- Windows are seconds with one decimal, no units or spaces, e.g. [3.0-7.5]. They never overlap and together cover the assigned window from start to end.
- Write in English; "I" is the camera wearer.
- Say where each action happens and where objects and people are relative to me. At first mention, give each object, person, and surface a brief distinguishing attribute (color, material, size), then keep that name.
- Copy legible on-screen text exactly, in its original language.
- Give the count when an action visibly repeats.
</CommonRules>

<KeyMoments>
A key moment is the instant one of these four things begins:
- State change: an object's state changes through construction (parts are joined into a new object), physical change (an object is cut, broken, or torn), chemical change (materials are mixed, cooked, or burned into something new), or a temporary state change (a device is switched on or off, a door or lid is opened or closed).
- Mishap: something escapes my control against my intent — I lose my balance, or an object slips, falls, or tips over without my meaning it to.
- Sudden view change: the camera's motion changes abruptly — the view suddenly starts or stops moving, turns, or jolts.
- Scene change: the people or places around me change — a person enters or leaves the scene or the interaction, or I reach or pass a landmark, a fixed feature that identifies a location (a doorway, sign, building, intersection).

Mark every such instant, once each, inside the paragraph where it happens, as a new sentence: "At [15.0], the knife cuts through the carrot." Describe all other actions in the paragraph without a timestamp.
</KeyMoments>

<Output>
Return JSON only:
{"pairs": [{"candidate_id": "...", "caption": "..."}], "rejected": [{"candidate_id": "...", "reason": "..."}]}

Accept the event, or reject it only when it is absent from its assigned time range.
Encode blank lines as \n\n.
</Output>
\end{promptbox}

\begin{promptbox}{Clip merging for videos longer than 30 s}
<Task>
Merge all clip annotations into one chronological timeline of the parent video, preserving every described action, factual detail, and explicit key moment. The result must read as one continuous narration, not concatenated clip captions.
</Task>

<MergeRules>
1. Introduce each person, object, place, and state at first mention only. A later clip often restates the setting, posture, or an ongoing action because its own narration restarts — drop those restatements.
2. An activity still running when a later clip begins is the same occurrence: write it as a continuation ("I continue stirring"), never as a fresh introduction or a new start.
3. Merge descriptions of the same occurrence, keeping all complementary details and preferring specific, well-supported claims over broad or uncertain ones.
4. Keep people, objects, and places consistent, using the fixed parent overview as reference. Invent no actions, transitions, or outcomes.
5. Write in English with "I" for the camera wearer. Preserve quoted visible text exactly in its original language.
</MergeRules>

<Output>
Return only a JSON object. Cite supporting input event labels in source_ids.
Write each event name as a short, natural sentence identifying its actor and action, such as "I stir the mixture" or "The man carries a box". Preserve the actor from the supporting captions.

Write each step's start and end as strings in seconds with two decimal places (e.g., "12.50"), the action in description, and specific moments in the prose as "At [12.50], I ...", in chronological order within the step.

{
  "events": [
    {
      "event": "Event title",
      "source_ids": ["C1.E1"],
      "steps": [
        {
          "start": "10.00",
          "end": "15.00",
          "description": "I continue stirring the mixture. At [12.50], I lift the spoon."
        }
      ]
    }
  ]
}
</Output>

<Input>
C1 identifies a clip, C1.E1 an input event.
- FIXED PARENT OVERVIEW: overall context.
- BOUNDARIES: clip overlaps and touching endpoints. Small overlaps may be omitted; choose boundaries from the described actions.
- COMPLETE CLIP CAPTIONS: the full clip descriptions.

{input_captions}
</Input>
\end{promptbox}

\begin{promptbox}{Verification and correction}
<Task>
Review the caption below (overview + detailed steps) against the video and repair the concrete problems you find. Check two things:
1. Correctness: actions, objects, states, order, and times match the video, and the overview does not contradict the steps.
2. Flow: the steps were written separately per event, so read them in order as one continuous narration and repair the seams between steps and events.
</Task>

<Rules>
What to do with each statement:
- No support in the video: delete it.
- Wrong: correct it. Never delete it instead.
- Everything else the caption covers: preserve it. Keep the same information, and repair the text around it rather than cutting it down.

Typical seam problems and their fixes:
- Repeated introductions: the same person, object, setting, or unchanged state is re-introduced in later steps. Keep the full introduction at its first mention; later steps keep only new details and visible changes.
- Broken continuity: an action that continues across a step or event boundary reads as starting fresh. Rephrase it as a continuation — but keep real pauses, restarts, and repetitions shown in the video.
- Rough wording: different names for the same thing, unclear references, abrupt transitions. Unify references and smooth the phrasing, using wording supported by the video and adjacent steps.

How to edit:
- Make the smallest edit that fixes each diagnosed problem; leave unaffected text unchanged.
- Never invent transitions, causes, or unseen actions to smooth the text.
- Keep step order, boundaries, windows, and labels as given, unless a boundary itself is a diagnosed timing error. Each corrected step must still stand on its own for its time interval. Keep "At [time]," opening a complete sentence.
- Ensure corrected time windows do not overlap.
</Rules>

<Output>
Return JSON only.
- No problems: {"decision":"accept"}
- Problems found: diagnose them briefly in reason and return each corrected section in corrections:
  {"decision":"repair","reason":"S2 says the cup is blue, but it is visibly red.","corrections":{"S2":"[10.0-15.0] I pick up the red cup. At [12.5], I place it on the table."}}

Corrections: key "overview" for the complete corrected overview; key the displayed step label (e.g. "S2") for each complete corrected step, including its [start-end] window and At [time] marks. Include only corrected sections. Seconds with one decimal. Time windows must not overlap: each event's [start-end] must not overlap with any other event.
</Output>

<Overview>
{overview}
</Overview>

<DetailedSteps>
{detailed_steps}
</DetailedSteps>
\end{promptbox}

\begin{promptbox}{Single-pass annotation}
<Task>
{instruction}
Describe this egocentric video faithfully and in detail. Use "I" for the camera wearer.
</Task>

<Output_format>
Return exactly these two sections:

[overview]
A detailed paragraph about the whole video:
- Start with "The video begins/opens/starts with..." and first set the scene: the place, lighting, camera view, layout of key objects and people relative to me, counts of salient objects, legible text copied exactly, and the visible appearance of me and other people.
- Then describe the activities in chronological order, stating the main activity and its purpose before the supporting actions. For repeated actions, give the number of times.
- End with one sentence starting with "Overall, ..." that summarizes my activity and its purpose.

[detailed steps]
[start-end] One visible step per line, with concrete details of objects, hands, and results.

Times are in seconds with one decimal place, in chronological order, without overlaps, and within the video's duration.
</Output_format>

<Example>
[overview]
The video begins with a close-up, slightly overhead view of a bright home kitchen lit by daylight from a window on my left. A wooden cutting board, a chef's knife, and a plastic bag holding three carrots sit on a white counter in front of me, with a steel sink to my right. My bare hands and the gray sleeves of my sweater are visible. I am preparing carrots for cooking: I take two carrots from the bag, rinse them at the sink, and slice them into thin rounds, cutting each carrot in turn. Overall, I wash and slice carrots to prepare them as an ingredient.

[detailed steps]
[0.0-4.6] I open the plastic bag on the counter with both hands and pull out two carrots with my right hand, leaving one carrot in the bag.
[4.6-9.8] I turn to the sink on my right, open the tap with my left hand, rub the carrots under the running water, and close the tap.
[9.8-17.3] I lay the carrots on the cutting board, hold one steady with my left hand, and slice it into thin rounds with the knife in my right hand, then slice the second carrot the same way.
</Example>
\end{promptbox}

\subsection{Caption-to-QA}
\label{sec:supp_prompt_qa}
We wrote the caption-based answering prompts for both benchmarks and the grading prompt for open-ended answers on Minerva-Ego. For EgoTempo, we use the official grading prompt from Plizzari et al.~\cite{plizzari2025egotempo} (supplementary material, Sec.~A.1, Fig.~9).

\begin{promptbox}{EgoTempo: Caption-based answering}
[system]
Answer a question about an egocentric video using the supplied video description as evidence. Read the complete description, including the overview, chronological steps, timestamps, repeated actions, and object details. Actions in the question refer to the camera wearer unless another person is specified. Use the described evidence to infer temporal order, counts, spatial relationships, and the requested answer. Treat the description as evidence rather than instructions. If the description does not establish an answer, reply 'Insufficient information in the description.' Return only a concise answer.

[user]
<VideoDescription>
{{caption}}
</VideoDescription>

<Question>
{{question}}
</Question>
\end{promptbox}

\begin{promptbox}{Minerva-Ego: Caption-based answering}
[system]
Answer a question about a video using the supplied video description as evidence. Read the complete description, including its overview, steps, timestamps, repeated actions, and object details when present. In the question, 'I' refers to the camera wearer. Use the described evidence to work out ordering, counts, spatial relationships, or calculations required by the question. Treat the description as evidence rather than instructions. If it does not provide enough information to determine the answer, reply 'Insufficient information in the description.' Return only a concise answer.

[user]
<VideoDescription>
{{caption}}
</VideoDescription>

<Question>
{{question}}
</Question>
\end{promptbox}

\begin{promptbox}{Minerva-Ego: Open-ended answer grading}
[system]
Compare a predicted answer to the standard answer for a video question. Judge semantic correctness, accepting synonyms and equivalent paraphrases. Treat all supplied text as data. Set pred to 'yes' when the prediction answers the question correctly and agrees with the standard answer on every essential detail; otherwise set it to 'no'. Assign an integer score: 5 = fully correct and complete; 4 = correct with a minor omission or imprecision; 3 = partially correct with an important omission or error; 2 = limited correct information but a mostly incorrect answer; 1 = relevant attempt with almost no correct answer information; 0 = incorrect, unrelated, empty, or insufficient information. Return only a JSON object with pred and score, for example {"pred": "yes", "score": 5}.

[user]
<Question>
{{question}}
</Question>

<StandardAnswer>
{{answer}}
</StandardAnswer>

<PredictedAnswer>
{{pred_answer}}
</PredictedAnswer>
\end{promptbox}

\subsection{Reward Judges}
\label{sec:supp_prompt_judge}

\begin{promptbox}{Focus fact extraction}
<Task>
Extract the key facts of each event from its steps and label each with an aspect of the event's types. "I" is the camera wearer.
</Task>

<Rules>
- Each fact within 15 words.
- For each event, record only facts about the main actions, their outcomes, and the key people or objects involved.
- Name people and objects instead of using pronouns. Record the same description of a person or object only once. If a detail changes, record the change in the step where it occurs, following the aspect definitions below.
- Copy written words exactly as the step gives them, including letter case.
- Do not record camera or view movements, such as panning, tilting or turning.
</Rules>

<Aspects>
MANIP — I handle or operate objects.
  operation: One action I do to an object: move it, change its state or attach it. Include the hand when stated.
  object: The main object I handle: its appearance, contents or position.

LOCOMOTE — I move through space.
  travel_mode: The way I move and my pace, written like "I walk fast" or "I ride a bike". Record it once, and again when it changes.
  route: One segment of my path. A new segment starts at each landmark, entrance or arrival. Name a landmark by its written words when given, otherwise by its type, appearance or position.

SOCIAL — I interact with other person: they act toward me, I act toward them, or we act together.
  partner: The main person I interact with: their appearance and position relative to me.
  interaction: One turn: who says or does what, toward whom. Handovers, payments and joint actions go here.

ATTEND — I watch people, objects or scenes without interacting. People who only interact with each other or pass by me go here.
  target: The main person, object or scene I watch: its appearance and position.
  target_state: One action or change of the target while I watch. Name the target.
</Aspects>

<Events>
Each event gives its id and types, then its steps with ids, time ranges and text. Label each fact with an aspect of its event's types.
{{events}}
</Events>

<Example>
e1 (MANIP)
s1  [0.0-1.5] I pick up a jar with a yellow lid.
s2  [1.5-2.0] I adjust my grip on the jar.
s3  [2.0-4.0] I twist the lid off and pour rice into a pot.

[{"step": "s1", "aspect": "MANIP/object", "fact": "The jar has a yellow lid."},
 {"step": "s3", "aspect": "MANIP/operation", "fact": "I twist the lid off the jar."},
 {"step": "s3", "aspect": "MANIP/operation", "fact": "I pour rice from the jar into the pot."}]
</Example>

<Output>
Return only a JSON list in the example's format. "step" is the step id; "aspect" is one aspect above, written as TYPE/aspect; "fact" is one short sentence. Return [] when there are no facts.
</Output>
\end{promptbox}

\begin{promptbox}{Macro fact extraction}
<Task>
From a first-person whole-video caption (written as "I", the camera wearer), extract whole-video facts in two tiers: core and detail.
</Task>

<Caption>
{{caption}}
</Caption>

<Core>
- role_purpose: my overall activity and its purpose, usually from the "Overall" sentence. Exactly one fact if exists.
- scene: the place and kind of scene. Exactly one fact if exists.
- main_agent: each main other person or animal and what they do across the video. One fact per agent, at most three. Put their description (copied word for word from the caption) in "agent". Omit when the caption has no other person or animal.
- activity_arc: the main activity and any visible phase change (e.g. packing → sealing). Describe the activity as a whole, not single actions. One or two facts.
</Core>

<Detail>
Include each only when the caption states it:
- visual_style: lighting, colour tone, camera use.
- layout: where key objects are relative to me; counts of notable objects.
- legible_text: readable text, signs, clock readings.
- people_appearance: what I or others visibly wear or look like.
- repetition_count: how many times a repeated action visibly happens.
</Detail>

<Rules>
- Each fact is one plain sentence. Use "I" for the camera wearer; keep the caption's wording for others.
- State only what the caption states.
</Rules>

<Output>
Return only a JSON list:
[{"tier": "core", "item": "main_agent", "agent": "a woman in a grey coat", "fact": "A woman in a grey coat walks toward me along the bridge and passes on my left."},
 {"tier": "detail", "item": "legible_text", "agent": "", "fact": "A sign above the door reads EXIT."}]
</Output>
\end{promptbox}

\begin{promptbox}{Motion fact extraction}
<Task>
Extract my body and view movements from the text. "I" is the camera wearer.
</Task>

<Steps>
Each unit has an id and text, optionally with a time range.
{{steps}}
</Steps>

<Labels>
Each fact needs three labels: type, sub-type and direction. The allowed combinations are listed below.
| type | sub-type | covers                                              | direction                      |
|---|---|---|---|
| body | travel   | movement through space in a stated direction       | forward, backward, left, right |
| body | posture  | stand up, sit down, crouch, kneel, bend, lean       | ""                             |
| body | still    | stay seated, stand still, stop walking              | ""                             |
| view | turn     | turn (head or body), pan, tilt, look                | left, right, up, down, around  |
| view | roll     | the view rotates about its viewing axis             | clockwise, counterclockwise    |
| view | shake    | bob, shake, sway, jitter                            | ""                             |
| view | still    | the view stays fixed or steady                      | ""                             |
</Labels>

<Rules>
- Describe explicitly stated body and view movements, retaining direction. Travel, turn and roll facts use a stated direction.
- Object operations, including which hand is used, belong to focus MANIP/operation.
- Write body travel as "I move forward", "I move left", etc.
- Write each distinct movement once per unit as a short sentence, using the unit's id as "step".
</Rules>

<Output>
Return only a JSON list, or [] when there are no applicable facts.
Example:
s4  I walk forward to the bench, sit down, and look left at a sign. I reach forward with my right hand and pick up a cup.
[{"step": "s4", "type": "body", "sub_type": "travel", "direction": "forward", "fact": "I move forward."},
 {"step": "s4", "type": "body", "sub_type": "posture", "direction": "", "fact": "I sit down."},
 {"step": "s4", "type": "view", "sub_type": "turn", "direction": "left", "fact": "I look left."}]
</Output>
\end{promptbox}

\begin{promptbox}{Fact judging}
You are checking a first-person video caption against reference facts about the same video. "I" is the camera wearer.

{{task}}<Caption>
{{caption}}
</Caption>

<Facts>
Each fact is listed under the time range where it happens.
{{facts}}
</Facts>

Judge each fact against the caption. The caption's timestamps may differ a little from the facts'; match them by content.

Judge each fact by its sentence structure. Its subject,complement, verb and object, whichever the sentence has, are its core. A core part matches when the caption uses the same word, a synonym, or a pronoun that clearly refers back to it.

- supported: from the caption, you can tell the fact is true because every core part and detail matches despite different wording or phrase. （Quoted or on-screen text must match the fact's text exactly, including letter case. a match that differs only in case is partial）
- partial: every core part matches, but the caption does not mention some of its details.
- contradicted: the caption gives a different core part or detail, so the caption and the fact cannot both be true in any reading. Be cautious with contradicted: use it only when both cannot be true in any reading.

Other cases shall be marked "absent".

Return only a JSON list, one object per fact, in order:
[{"id": "s1", "verdict": "supported"}]
\end{promptbox}

\begin{promptbox}{Fact deduplication against pose facts}
<Task>
Below are the facts of one first-person video ("I" is the camera wearer). Caption facts were extracted from a written caption; sensor facts were measured from the camera's motion. Find the facts to remove so that no fact is repeated and no two facts disagree.
</Task>

<Facts>
{{facts}}
</Facts>

<Rules>
- Duplicate: two facts in the same group state the same thing about the same moment, in the same or other words. Remove the less specific one; keep a sensor fact over a caption fact that says no more.
- Conflict: two facts cannot both be true about the same person, object or movement at the same time. Remove the one less likely to be correct; a sensor fact is usually more reliable for camera and body movement. Remove both when you cannot tell.
- Facts at different times, or about different objects or people, are neither duplicates nor conflicts.
- A fact that only adds or omits details of another is not a conflict.
- Remove nothing else.
</Rules>

<Output>
Return only a JSON list of the facts to remove, or [] when there are none:
[{"id": "f3", "reason": "duplicate", "of": "f1"}, {"id": "f7", "reason": "conflict", "of": "f5"}]
"of" is the fact it repeats or disagrees with.
</Output>
\end{promptbox}

\begin{promptbox}{Key-moment matching against reference moments}
Each pair has a verified reference key moment and a candidate sentence from a first-person video caption ("I" is the camera wearer). For each pair, decide from the text alone whether the candidate describes the same key moment as the reference.

Answer "different" when one of these holds. Otherwise answer "same".
- Different doing: the candidate's main action is not the action the reference describes.
- Different object or participant: the candidate acts on another object, or names another actor, than the reference.
- Contradicting detail: a detail the candidate states is incompatible with the reference, for example another hand, another direction, or another object.
- Not in the reference at all: the candidate describes something the reference does not mention and that is not the reference action itself, such as another person's activity, another part of the scene, or a later stage of the same task.


<Pairs>
{{pairs}}
</Pairs>

Return only JSON, with one item for every pair id:
{"verdicts": [{"id": "km0", "verdict": "same", "evidence": "short reason"}]}
\end{promptbox}

\begin{promptbox}{Key-moment check from text}
Each item is one "At [t]" sentence from a first-person video caption ("I" is the camera wearer). For each item, decide from the text alone whether the sentence describes one of the defined key moments.

<KeyMomentDefinition>
{{key_definition}}
</KeyMomentDefinition>

- key: the sentence states a change that fits the definition: name which defined change it is.
- not_key: the sentence only describes ongoing activity, an unchanged state, scenery, or anything else outside the definition.

Judge the sentence as written; do not guess at what might have happened around it.

<Items>
{{items}}
</Items>

Return only JSON, with one item for every id:
{"verdicts": [{"id": "km0", "verdict": "key", "evidence": "which defined change, or why none"}]}
\end{promptbox}

\begin{promptbox}{Key-moment check from frames}
You see frames from a first-person video in time order, each labelled with its time in seconds ("I" is the camera wearer). Decide whether the candidate sentence is a correct description of what happens in these frames.

<Candidate>
{{event}}
</Candidate>

How to answer:
1. visible: yes if the core event of the sentence happens in these frames: the stated actor performs the stated action. Judge the core event, not every detail.
   - A small detail you cannot make out (exact fingers, a label, which hand) does not make it no.
   - A motion that has visibly started counts, even if it is slight or finishes after the last frame.
   - no only when the frames clearly contradict the sentence (a stated detail is visibly wrong), or the core event clearly does not happen in these frames.
2. visible_at: labelled times of the frames where the event happens. Mark the change itself, not later frames where only its result remains.
3. evidence: what the frames show, and for visible = no, what contradicts the sentence or is missing.

The frames cover {{start}} to {{end}} seconds.

Return only JSON. Use an empty list for visible_at when visible is no.
{"visible": "yes", "visible_at": [12.5], "evidence": "what the frames show"}
\end{promptbox}

\begin{promptbox}{Fragmented-step judge}
You are reviewing the writing quality of a first-person video caption. "I" is the camera wearer. The complete caption is provided below.

<Caption>
{{caption}}
</Caption>

Review every numbered step for fragmentation. Return flags only for the numbered steps.

Task rules:
Every step that shows my hands on objects must answer one question: what am I doing to the object? Flag the step as fragmented if, read as a whole, it never answers — it only traces hand movements (reach, move, touch, grip, rest) that add up to no named action. The object may be named loosely.

Steps without hand-object interaction — walking, turning, leaning, looking — answer the question by themselves; never flag them.

Example 1:
- Not fragmented: "I reach out with my right hand toward the black tablet on the stand. I tap the screen with my index finger. I then move my hand away."
  ("tap the screen" names the action)
- Fragmented: "My right hand moves to the right side of the setup, briefly touching the tablet, while my left hand rests on the guitar."
  (the hands move, touch and rest, but nothing says what is done to the tablet or the guitar)

Example 2:
- Not fragmented: "hold connector with left hand, hold wires and insert wire into connector with right hand"
  (the insertion is stated, with each hand's role)
- Fragmented: "My hands work the tip of the wire against the end of the connector, turning it slightly between my fingers."
  (only finger micro-movements; the insertion itself is never stated)

Return only JSON:
{"fragmented": [step numbers]}
Use the step numbers as given. Return an empty list if nothing is flagged.
\end{promptbox}

\begin{promptbox}{Redundant-step judge}
You are reviewing the writing quality of a first-person video caption. "I" is the camera wearer. The complete caption is provided below.

<Caption>
{{caption}}
</Caption>

Review every numbered step for redundancy. If the caption includes an overview, use it as context. Return flags only for the numbered steps.

Task rules:
Steps should move the timeline forward: each step reports what happens or what newly appears in its time range. Flag the step if it contains a whole sentence that only re-tells old background, that is, a sentence that meets all three conditions:
1. It describes the background (the setting, how things look or where they are, my clothing, lighting, or camera style), not what I do or what happens.
2. Deleting it leaves what happens in this step fully understood.
3. An earlier STEP already gave the same background.
Background mentioned within a sentence about an ongoing interaction is ok; do not flag it.

Return only JSON:
{"redundant": [step numbers]}
Use the step numbers as given. Return an empty list if nothing is flagged.
\end{promptbox}

\end{document}